\documentclass[10pt,twocolumn,letterpaper]{article}

\usepackage{cvpr}
\usepackage{times}
\usepackage{epsfig}
\usepackage{graphicx}
\usepackage{amsmath}
\usepackage{amssymb}

\usepackage{booktabs}
\usepackage{multirow}
\usepackage{bm}
\usepackage{subfigure}
\usepackage{color}
\usepackage[table]{xcolor}
\usepackage{bbm}
\usepackage[breaklinks=true,bookmarks=false]{hyperref}

\cvprfinalcopy % *** Uncomment this line for the final submission

\def\cvprPaperID{****} % *** Enter the CVPR Paper ID here

\begin{document}

%%%%%%%%% TITLE
\title{Unified Loss of Pair Similarity Optimization for Vision-Language Retrieval}

%\author{Zheng Li$^{1}$\and
%	Caili Guo$^{1,2}$\footnote{Corresponding author}\and
%	Zerun Feng$^{1}$\and
%	Jenq-Neng Hwang$^3$\and
%	Xijun Xue$^4$
%	\affiliations
%	$^1$Beijing Key Laboratory of Network System Architecture and Convergence, \\Beijing University of Posts and Telecommunications\\
%	$^2$Beijing Laboratory of Advanced Information Networks\\
%	$^3$University of Washington\\
%	$^4$China Telecom System Integration Co.,Ltd\\
%	\emails
%	\{lizhengzachary, guocaili, fengzerun\}@bupt.edu.cn,
%	hwang@uw.edu,
%	xuexj@chinatelecom.cn
%}

\author{Zheng Li$^{1}$, Caili Guo$^{1}$, Xin Wang$^{1}$, Zerun Feng$^{1}$, Jenq-Neng Hwang$^{2}$, Zhongtian Du$^{3}$ \\
	$^1$Beijing University of Posts and Telecommunications \\
	$^2$University of Washington \\
	$^3$China Telecom Digital Intelligence Technology Co., Ltd. \\
	{\tt\small 
		\{lizhengzachary, guocaili, wangxin1999, fengzerun\}@bupt.edu.cn,} \\
	{\tt\small hwang@uw.edu, duzt@chinatelecom.cn}
}

\maketitle
%\thispagestyle{empty}

%%%%%%%%% ABSTRACT
\begin{abstract}
There are two popular loss functions used for vision-language retrieval, \textit{i.e.},  triplet loss and contrastive learning loss, both of them essentially minimize the difference between the similarities of negative pairs and positive pairs. More specifically, Triplet loss with Hard Negative mining (Triplet-HN), which is widely used in existing retrieval models to improve the discriminative ability,  is easy to fall into local minima in training. On the other hand, Vision-Language Contrastive learning loss (VLC),  which is widely used in the vision-language pre-training, has been shown to achieve significant performance gains on vision-language retrieval, but the performance of fine-tuning with VLC on small datasets is not satisfactory. This paper proposes a unified loss of pair similarity optimization for vision-language retrieval, providing a powerful tool for understanding existing loss functions.  Our unified loss includes the hard sample mining strategy of VLC and introduces the margin used by the triplet loss for better similarity separation. It is shown that both Triplet-HN and VLC are special forms of our unified loss. Compared with the Triplet-HN, our unified loss has a fast convergence speed. Compared with the VLC, our unified loss is more discriminative and can provide better generalization in downstream fine-tuning tasks. Experiments on image-text and video-text retrieval benchmarks show that our unified loss can significantly improve the performance of the state-of-the-art retrieval models.
\end{abstract}

%%%%%%%%% BODY TEXT
\section{Introduction}
Vision-language retrieval, such as image-text retrieval \cite{cheng2022vista, zhang2022negative, zhang2022show} and video-text retrieval \cite{shvetsova2022everything, ge2022bridging, gorti2022x, cao2022visual, wang2022object}, \textit{etc.}, is formulated to retrieve relevant samples across different vision and language modalities.
Compared to unimodal image retrieval, vision-language retrieval is more challenging due to the heterogeneous gap between query and candidates.

The mainstream approach to achieve vision-language retrieval is to learn a shared visual-semantic embedding space, where the similarity between the embedding vectors of related vision and language instances is maximized and the irrelevant is minimized.
Existing loss functions are mainly based on the optimization of the similarity of sample pairs, in which related samples are called positive pairs, and irrelevant are called negative pairs. 
A variety of loss functions have been proposed for vision-language retrieval \cite{faghri2018vse++, wei2020universal}.
Two popular loss functions among them are triplet loss and contrastive learning loss.

Triplet loss with online Hard Negative mining (Triplet-HN) \cite{faghri2018vse++} is the most widely used loss function in vision-language retrieval.
The optimization objective of Triplet-HN is to make the similarity of positive pairs greater than negative pairs.
Triplet-HN incorporates hard negatives in the loss function, which yields significant gains in retrieval performance. 
Many loss functions \cite{wei2020universal, wei2021universal} proposed recently are variants of it. 
However, Triplet-HN only mines the negative pair with the largest similarity in each batch, which makes the model easy to fall into local minima, resulting in bad convergence.

In recent years, vision-language pre-training models \cite{li2021align} have been intensively explored to bridge vision and language. 
The paradigm of vision-language pre-training is to design pre-training tasks on large-scale vision-language data for pre-training and then fine-tune the model on specific downstream tasks.
These models, which have achieved large performance gains on vision-language retrieval, use a Vision-Language Contrastive learning loss (VLC) to learn whether vision and language samples match or not.
We find that VLC is essentially minimizing the difference between the similarities of negative and positive pairs.
From the perspective of pair similarity optimization, it can be seen that VLC has insufficient discriminative for positive and negative samples, which affects the generalization of the model, especially the performance of fine-tuning on small datasets.

This paper proposes a unified loss of pair similarity optimization for vision-language retrieval, providing a powerful tool for understanding existing loss functions. 
More specifically, we convert VLC into a form of pair similarity optimization and find that VLC is insufficient in distinguishing positive and negative samples. 
By introducing the margin used by the triplet loss to VLC, we unify these two classes of loss functions.
We further show that most of the existing loss functions are variants of the unified loss. 
By analyzing the gradient of the loss functions, we explain the reason why Triplet-HN is easy to fall into local minima and prove that our unified loss has a faster convergence speed.
Besides, we demonstrate that our unified loss can provide better generalization in fine-tuning compared to VLC.
The major contributions of this paper are summarized as follows:
\begin{itemize}
	\item A unified loss of pair similarity optimization is proposed for vision-language retrieval.
	Most existing loss functions are shown to be variants of the unified loss.
	\item Our theoretical analysis and experiments verify that the unified loss has a faster convergence speed and better generalization compared with existing loss functions.
	\item Experiments on image-text and video-text retrieval benchmarks demonstrate that our unified loss can significantly improve the performance of state-of-the-art vision-language retrieval models.
\end{itemize}

\section{Related Work}
\subsection{Image-Text Retrieval} 
Image-Text Retrieval (ITR) is the main application of vision-language retrieval.
Existing ITR methods can be divided into two categories according to the image-text matching methods, \textit{i.e.}, global-level matching methods \cite{li2019visual, wang2020consensus, chen2021learning} and local-level matching methods \cite{chen2020imram, liu2020graph, zhang2020context}.
The global-level matching methods embed the whole images and sentences into a joint embedding space, and the matching score between the embeddings of images and sentences can be calculated by a simple similarity metric (\textit{e.g.} cosine similarity).
\cite{frome2013devise} propose the first global-level matching model DeViSE, which employs the CNN and Skip-Gram to project images and sentences into a joint embedding space.
The local-level matching methods obtain the matching score by calculating the cross-attention between image region and words. 
\cite{lee2018stacked} propose a Stacked Cross Attention Network (SCAN), which measures the image-text similarity by aligning image regions and words.

\subsection{Video-Text Retrieval}
Video-Text Retrieval (VTR), as an important branch of vision-language retrieval, has received extensive attention in recent years. 
Compared to images, videos contain more complex temporal information. 
Many VTR works are devoted to learning rich video semantic representations.
\cite{dong2019dual} encode videos and texts by a CNN and a Bi-GRU, and employ mean pooling to get multi-levels representations.
Several works \cite{liu2019use} introduce the multi-modal features extracted from videos for retrieval, such as motion and audio features.
\cite{gabeur2020multi} present a multi-modal transformer to jointly encode the different modalities in videos for retrieval.
While above ITR and VTR methods learn an advanced encoding network to generate richer semantic representations. 
They mainly adopt Triplet-HN in the optimization. 
Our proposed unified loss is independent of the encoding network of these retrieval models. 
It can directly replace the original loss function to improve the performance of the retrieval model.

\subsection{Deep Metric Learning}
The main work of this paper belongs to the field of Deep Metric Learning (DML). 
DML aims to construct an embedding space to reflect the semantic distances among instances. 
It has many applications such as image retrieval \cite{oh2016deep} and face recognition \cite{schroff2015facenet}. 
Triplet loss \cite{hoffer2015deep} is a representative method in DML, which aims to force the similarities of positive pairs to be higher than that of negative pairs by a margin.
Several works \cite{wang2019multi, sun2020circle} provide a unified perspective on DML.
\cite{wang2019multi} establish a general pair weighting framework for pair-based loss functions and propose a multi-similarity loss.
\cite{sun2020circle} provide a unified perspective of pair similarity optimization and propose a circle loss.
The above methods are all aimed at unimodal image retrieval, while cross-modal DML has its own characteristics. 
In vision-language retrieval, each anchor has only one corresponding positive sample, which makes many methods of unimodal DML not directly applicable to cross-modal settings.

\subsection{Deep Metric Learning for Vision-Language Retrieval} 
DML for vision-language retrieval is also a significant research field.
In recent years, a variety of DML methods have been proposed \cite{faghri2018vse++, wei2020universal, chen2020adaptive}. 
A hinge-based triplet loss is widely used as an objective to force positive pairs to have higher similarity scores than negative pairs by a margin \cite{frome2013devise}. 
\cite{faghri2018vse++} incorporate hard negatives in the triplet loss, which yields significant gains in retrieval performance. 
There are a number of works \cite{wei2020universal, chen2020interclass, wei2021universal} that propose weighting DML frameworks for vision-language retrieval, which can further improve retrieval performance. 
Although the forms of various loss functions are quite different, most of them are variants of Triplet-HN. 
Triplet-HN only mines the hardest negative pair in each batch, which will make the model easy to fall into local minima and affect the convergence of training. 
Triplet-HN is a special forms of our unified loss. 
Compared with Triplet-HN, our unified loss incorporates more negative pairs into training and has fast convergence speed.

\subsection{Vision-Language Pre-training}
In recent years, Vision-Language Pre-training (VLP) \cite{chen2020uniter, radford2021learning, jia2021scaling} have been intensively explored to bridge vision and language. 
The paradigm of VLP is to design pre-training tasks on large-scale vision-language data for pre-training and then fine-tune the model on specific downstream tasks.
To take advantage of extensive natural language supervision information, CLIP \cite{radford2021learning} predicts which sentence matches which image, resulting in a transferable visual model.
ALIGN \cite{jia2021scaling} further extends CLIP with a dataset of over one billion image-text pairs.
These models have achieved large performance gains on vision-language retrieval. 
They use a VLC to learn whether images and sentences match or not.
We convert VLC into a form of pair similarity optimization and find that VLC has insufficient discriminative ability for positive and negative samples, which affects the generalization ability of the model.
VLC is also a special forms of our unified loss.
Compared with the VLC, our unified loss function is discriminative and can provide better generalization.

\section{Unified Loss of Pair Similarity Optimization}
\subsection{Preliminaries} \label{Preliminaries}
We first introduce the related background of vision-language retrieval. 
Given a set of visual instances $ \bm{V} $ and a corresponding set of sentences $ \bm{T} $. 
Let $ V_{i} $ be a visual instance, $ T_{i} $ be a sentence, $ {(V_{i}, T_{i})} $ be a vision-language instance pair, where components of the pair come from different modalities. 
In the case of image-text retrieval, given a sentence $ T_{i} $, the goal is to find the most relevant image $ V_{i} $ in the image gallery. 
The mainstream approach to this task is to learn a shared visual-semantic embedding space, where the similarity between the embedding vectors of related vision and language instances is maximized and the irrelevant is minimized.
The core idea behind the method is that there exists a mapping function, $ s(V_{i}, T_{i}; \bm{W}) = \bm{\Phi}(V_{i})^{\top} \bm{W} \bm{\Psi}(T_{i}) $  to measure the similarity score between the visual features $ \bm{\Phi}(V_{i}) $ and the text features $ \bm{\Psi}(T_{i}) $, where $ \bm{W} $ denote the parameters of $ s(\cdotp, \cdotp) $.
Loss functions for visual-language retrieval can be mainly divided into two categories: triplet loss and contrastive learning loss.

\textbf{Triplet loss with online Hard Negative mining (Triplet-HN)} is the most widely used loss function in visual-language retrieval. 
Triplet-HN takes the form of:
\begin{equation} \label{triplet}
	\begin{aligned}
		\mathcal{L}_{\text{Triplet-HN}}
		= & \sum_{i=1}^{B}
		\left(
		\left[ 
		s(V_{i}, \hat{T}_{i}) - s(V_{i}, T_{i}) + m 
		\right]_{+}
		\right. \\
		& \left. 
		+ \left[ 
		s(\hat{V}_{i}, T_{i}) - s(V_{i}, T_{i}) + m 
		\right]_{+}
		\right),
	\end{aligned}
\end{equation}
where 
\begin{equation}
	\begin{aligned}
		\hat{T}_{i} = \max_{j=1, i \neq j}^{B} s{(V_{i}, T_{j})}, \\
		\hat{V}_{i} = \max_{j=1, i \neq j}^{B} s{(V_{j}, T_{i})}.
	\end{aligned}
\end{equation}
$ B $ is the batch size, $ m $ is a margin for better similarity separation, $ [x]_{+} = \max (x, 0) $, 
$ \hat{T}_{i} $ and $ \hat{V}_{i} $ are the hard negative samples.
Triplet-HN incorporates the hardest negative in triplet loss, which yields significant gains in retrieval performance.

\textbf{Vision-Language Contrastive learning loss (VLC)} is widely used in visual-language pre-training. 
To simplify the representation, we abbreviate the similarity $ s{(V_{i}, T_{i})} $ of positive pairs $ (V_{i}, T_{i}) $ as $ s_{ii} $, and the similarity $ s{(V_{i}, T_{j})} $ of negative pairs $ (V_{i}, T_{j}) $ as $ s_{ij} $. 
VLC takes the form of:
\begin{equation}
	\mathcal{L}_{\text{VLC}}
	= - \sum_{i=1}^{B}
	\left(
	\log 
	\dfrac{e^{\gamma s_{ii}}}
	{\sum_{j=1}^{B} e^{\gamma s_{ij}}}
	+ \log 
	\dfrac{e^{\gamma s_{ii}}}
	{\sum_{j=1}^{B} e^{\gamma s_{ji}}} \right),
\end{equation}
where $ \gamma $ is a scale factor to control the hardness of mining hard samples.
Pre-training models use VLC to learn whether vision and language samples match or not, which have achieved large performance gains.

\begin{figure}[t]
	\centering
	\includegraphics[width=\linewidth]{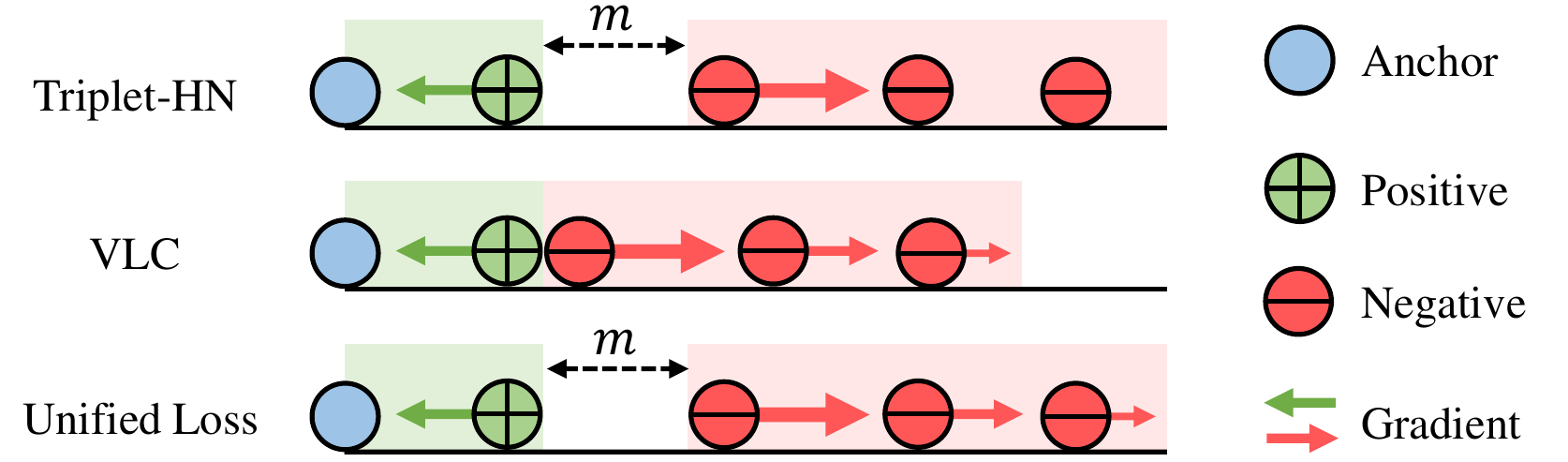}
	\caption{
		The direction of the arrow represents the gradient direction brought by loss functions during backpropagation. 
		The length and thickness of the arrows represent the magnitude of the gradient.}
	\label{loss}
\end{figure}
\subsection{Unified Loss}
From Eq.~\eqref{triplet} we can easily see that the optimization goal of $ \mathcal{L}_{\text{Triplet-HN}} $ is to make the differences $ (s(V_{i}, T_{i}) - s(V_{i}, \hat{T}_{i})) $ and $ (s(V_{i}, T_{i}) - s(\hat{V}_{i}, T_{i})) $ greater than a margin $ m $.
As \figurename~\ref{loss} shows, Triplet-HN only mines the negative sample pair with the largest similarity in each batch to participate in training, which could easily make the model easy to fall into local minima and affect the convergence of training. 

Although $ \mathcal{L}_{\text{VLC}} $ formally optimizes the ratio between the similarity of the positive sample pair and the similarities of all sample pairs.
VLC can also be converted into a form that optimizes pair similarity:
\begin{equation} \label{vlc}
	\begin{aligned}
		\mathcal{L}_{\text{VLC}}
		= & \sum_{i=1}^{B}
		\left(
		\log
		\left( 1 + \sum_{j=1, i \neq j}^{B} 
		e^{ 
			\gamma \left( s_{ij} - s_{ii} 
			\right)} \right) \right. \\
		& + \left. \log 
		\left( 1 + \sum_{j=1, i \neq j}^{B} 
		e^{
			\gamma \left( s_{ji} - s_{ii} \right)
		} \right) \right).
	\end{aligned}
\end{equation}
Please refer to the Appendix for the derivation of the equations involved in this paper.
From Eq.~\eqref{vlc} we find that $ \mathcal{L}_{\text{VLC}} $ is essentially optimizing the differences $ (s_{ij} - s_{ii}) $ and $ (s_{ji} - s_{ii}) $.
As \figurename~\ref{loss} shows, the optimization goal of $ \mathcal{L}_{\text{VLC}} $ is to make $ s_{ii} > s_{ij} $ and $ s_{ii} > s_{ji} $, which does not leave enough margin for the distinction between positive and negative samples.
Therefore, $ \mathcal{L}_{\text{VLC}} $ has insufficient discriminative ability for positive and negative samples, which affects the generalization ability of the model, especially the performance of fine-tuning on small datasets.

Both of the above loss functions have their own drawbacks. 
We integrate the above two loss functions, introducing the margin $ m $ used in the triplet loss into VLC, and propose a unified loss:
\begin{equation}
	\label{unified}
	\begin{aligned}
		\mathcal{L}_{\text{Unified}}
		= \frac{1}{\gamma} \sum_{i=1}^{B}
		\left(
		\log
		\left( 1 + \sum_{j=1, i \neq j}^{B} 
		e^{ 
			\gamma \left( s_{ij} - s_{ii} + m
			\right)} \right) \right. \\
		+ \left. \log 
		\left( 1 + \sum_{j=1, i \neq j}^{B} 
		e^{
			\gamma \left( s_{ji} - s_{ii} + m \right)
		} \right) \right).
	\end{aligned}
\end{equation}
As can be seen from Eq.~\eqref{unified}, the optimization goal of $ \mathcal{L}_{\text{Unified}} $ is to make the differences $ (s_{ii} - s_{ij}) $ and $ (s_{ii} - s_{ji}) $ greater than a margin $ m $.
The loss functions used for vision-language retrieval have various forms, and it is not easy to analyze the optimization direction of each loss function. 
Our unified loss provides a powerful tool for understanding existing loss functions.
Most existing loss functions are variants of $ \mathcal{L}_{\text{Unified}} $. 
Although their forms are different, their optimization goals are essentially the same.

Widely used $ \mathcal{L}_{\text{Triplet-HN}} $ is a special forms of $ \mathcal{L}_{\text{Unified}} $. 
When $ \gamma \rightarrow +\infty $, $ \mathcal{L}_{\text{Unified}} $ is transformed into $ \mathcal{L}_{\text{Triplet-HN}} $:
\begin{equation}
	\mathcal{L}_{\text{Triplet-HN}}
	= \lim_{\gamma \rightarrow +\infty}
	\mathcal{L}_{\text{Unified}}, 
\end{equation}
where $ \gamma \rightarrow +\infty $ represents that the loss only mines the hardest negative sample in a batch.
Our unified loss uses $ \gamma $ to control the hardness of mining samples, introducing more negative samples in training. 
Compared with Triplet-HN, the unified loss converges faster.
In Subsection \ref{analysis}, we theoretically analyze the reasons why Triplet-HN is easy to fall into local minima, and why our unified loss can have better convergence.
Moreover, $ \mathcal{L}_{\text{VLC}} $ is also a special forms of our unified loss. 
When $ m = 0 $, $ \mathcal{L}_{\text{Unified}} $ is transformed into $ \mathcal{L}_{\text{VLC}} $:
\begin{equation}
	\mathcal{L}_{\text{VLC}}
	= \gamma \cdot
	\mathcal{L}_{\text{Unified}}(m=0),
\end{equation} 
Our unified loss introduces a margin $ m $ for better similarity separation, which can improve the generalization of the retrieval model. 
There are a number of works \cite{wei2020universal, chen2020interclass, wei2021universal} that propose weighting metric learning frameworks for vision-language retrieval, which can further improve retrieval performance. 
They can also be included in our unified loss, but the pair similarities are multiplied by the weights.
The conversion between more loss functions and our unified loss are listed in the Appendix.

Compared to other loss functions, $ \mathcal{L}_{\text{Unified}} $ has a simple form, which does not require complex sampling strategies, and has the same complexity of $ O(N^{2}) $ like $ \mathcal{L}_{\text{Triplet-HN}} $ and $ \mathcal{L}_{\text{VLC}} $. 
$ \mathcal{L}_{\text{Unified}} $ can be plug-and-played into existing vision-language retrieval models, just by replacing the original loss.
Subsequent experiments show that using $ \mathcal{L}_{\text{Unified}} $ can improve the performance of existing visual-language retrieval models. 
Using a simple loss function $ \mathcal{L}_{\text{Unified}} $ can outperform other loss functions with complex sampling strategies.

\begin{figure}[t]
	\centering
	\includegraphics[width=\linewidth]{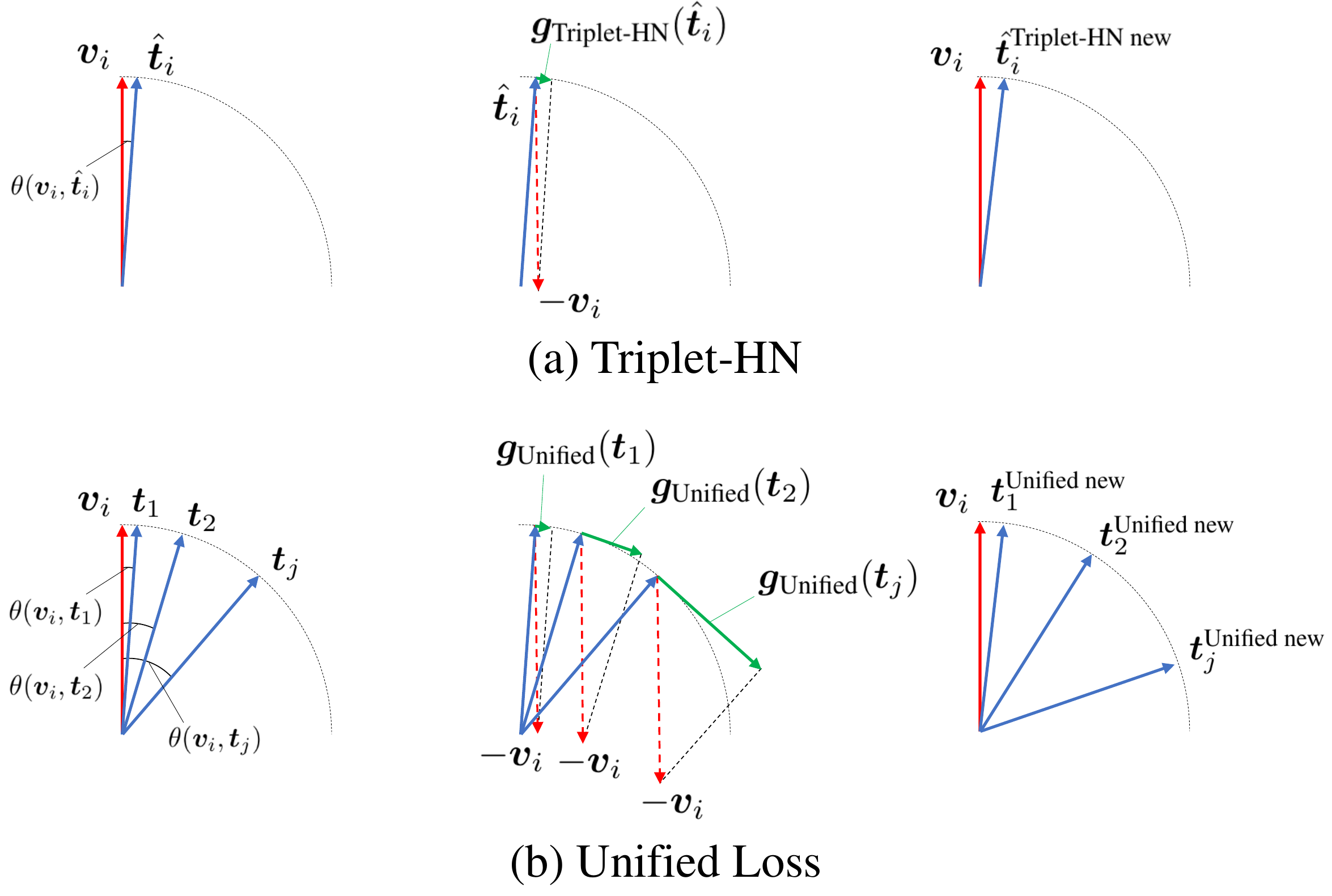}
	\caption{
		The red and blue arrows represent the feature vectors of the anchor and negative samples, respectively.
		The green arrows represent the components of the gradient of the loss function in the direction of the hypersphere tangent.}
	\label{gradient}
\end{figure}
\subsection{Gradient Analysis} \label{analysis}
In this section, we analyze the gradient of Triplet-HN, VLC and our unified loss on the learning of neural networks, and prove that unified loss can achieve better convergence and generalization.  
We consider the neural networks $ \bm{\Phi}(\cdot) $ and $ \bm{\Psi}(\cdot) $ embed the image and text on a unit hypersphere. 
We use $ (\bm{v}_{i}, \bm{t}_{i}) $ to simplify the representation of the normalized feature vectors. 
When embedded on a unit hypersphere, the cosine similarity is a convenient metric to measure the similarity between image-text pair $ s(V_{i}, T_{i}) = \bm{v}_{i}^{\top} \bm{t}_{i} $, and this similarity is bounded in the range $ [-1, 1] $. 

To simplify the representation, we only analyze the loss for vision-to-language retrieval, which is symmetric to the loss from language-to-vision retrieval.
Triplet-HN for vision-to-language retrieval can be written as:
\begin{equation}
	\mathcal{L}_{\text{Triplet-HN}}
	(\bm{v}_{i})
	= \max \left( 
	\bm{v}_{i}^{\top} \hat{\bm{t}}_{i} - \bm{v}_{i}^{\top} \bm{t}_{i} + m, 0
	\right).
\end{equation}
We derive the loss gradient with respect to the feature vectors $ \bm{v}_{i} $ and $ \hat{\bm{t}}_{i} $:
\begin{equation}
	\frac{
		\partial 
		\mathcal{L}_{\text{Triplet-HN}}
		(\bm{v}_{i})}
	{\partial \bm{v}_{i}} 
	= ( \hat{\bm{t}}_{i} - \bm{t}_{i} )
	\cdot \mathbbm{1}
	(\mathcal{L}_{\text{Triplet-HN}}
	(\bm{v}_{i}) > 0 ),
\end{equation}
\begin{equation}
	\frac{
		\partial 
		\mathcal{L}_{\text{Triplet-HN}}
		(\bm{v}_{i})}
	{\partial \hat{\bm{t}}_{i}} 
	= \bm{v}_{i}
	\cdot \mathbbm{1}
	(\mathcal{L}_{\text{Triplet-HN}}
	(\bm{v}_{i}) > 0 ),
\end{equation}
where if $ x $ is true, $ \mathbbm{1} \{ x \} = 1 $, otherwise $ \mathbbm{1} \{ x \} = 0 $.
The gradient direction of the loss function is related to the direction of the feature vector, $ \bm{v}_{i} $, $ \hat{\bm{t}}_{i} $. 
The gradient can be decomposed into two components. 
One component of the hypersphere tangent direction that contributes to the optimization of the loss function, such as the gradient component indicated by the green arrow in \figurename~\ref{gradient}.
The other is the component perpendicular to the tangent direction of the hypersphere, which pushes the features away from the hypersphere, and after normalization it does not work for the optimization of the loss function.
We use $ \theta(\cdot, \cdot) $ to represent the angle between the two feature vectors, and the modulo value of the component of the gradient in the direction of the hypersphere tangent can be expressed as:
\begin{equation}
	\left \Vert 
	\bm{g}_{\text{Triplet-HN}}(\bm{v}_{i}) 
	\right \Vert 
	=
	\left \Vert
	\frac{
		\partial 
		\mathcal{L}_{\text{Triplet-HN}}
		(\bm{v}_{i})}
	{\partial \bm{v}_{i}} 
	\right \Vert
	\cdot \sin 
	\theta(\bm{v}_{i}, \hat{\bm{t}}_{i} - \bm{t}_{i}),
\end{equation}
\begin{equation}
	\left \Vert 
	\bm{g}_{\text{Triplet-HN}}(\hat{\bm{t}}_{i}) \right \Vert 
	=
	\left \Vert 
	\frac{
		\partial 
		\mathcal{L}_{\text{Triplet-HN}}
		(\bm{v}_{i})}
	{\partial \hat{\bm{t}}_{i}} 
	\right \Vert 
	\cdot \sin 
	\theta(\bm{v}_{i}, \hat{\bm{t}}_{i}).
\end{equation}
We can use this gradient to update the features toward the decreased loss.
$ \hat{\bm{t}}_{i} $ is the hardest negative sample that Triplet-HN mines in each batch. 
In other words, it is the negative sample with the greatest similarity to anchor $ \bm{v}_{i} $.
In the beginning of training, it is easy to appear that the similarity between $ \hat{\bm{t}}_{i} $ and $ \bm{v}_{i} $ is very large.
When $ \bm{v}_{i}^{\top} \hat{\bm{t}}_{i} \rightarrow 1 $, $ \sin 
\theta(\bm{v}_{i}, \hat{\bm{t}}_{i}) \rightarrow 0 $, $ \Vert \bm{g}_{\text{Triplet-HN}}(\hat{\bm{t}}_{i}) \Vert \rightarrow 0 $, the gradient component that works for $ \hat{\bm{t}}_{i} $ optimization will approach $ 0 $.
Besides, the similarity between $ \hat{\bm{t}}_{i} $ and $ \bm{t}_{i} $ is also easy to be large in the initial stage of training.
When $ \bm{t}_{i}^{\top} \hat{\bm{t}}_{i} \rightarrow 1 $, $ \Vert \hat{\bm{t}}_{i} - \bm{t}_{i} \Vert \rightarrow 0 $, $ \Vert \bm{g}_{\text{Triplet-HN}}(\bm{v}_{i}) \Vert \rightarrow 0 $, the gradient component that works for $ \bm{v}_{i} $ optimization will approach $ 0 $.
As shown in \figurename~\ref{gradient}, after the gradient is updated, the updated feature vector $ \hat{\bm{t}}_{i}^{\text{Triplet-HN new}} $ does not change much compared to the original feature vector $ \hat{\bm{t}}_{i} $.
Therefore, in the early stage of training, Triplet-HN is easily suffered by the vanishing gradient.
This is the reason why Triplet-HN is easy to fall into local minima and affect the convergence of training. 

Our unified loss does not have the above problems during training.
To simplify the representation, we set $ \gamma $ to $ 1 $.
We also write the unified loss for vision-to-language retrieval:
\begin{equation}
	\mathcal{L}_{\text{Unified}}
	(\bm{v}_{i})
	= \log
	\left( 
	1 + \sum_{j=1, i \neq j}^{B} 
	e^{  
		\left( \bm{v}_{i}^{\top} \bm{t}_{j}
		- \bm{v}_{i}^{\top} \bm{t}_{i} + m
		\right)} 
	\right).
\end{equation}
The loss gradient with respect to the feature vectors $ \bm{v}_{i} $ and $ \bm{t}_{j} $ can be written as:
\begin{equation}
	\frac{
		\partial 
		\mathcal{L}_{\text{Unified}}
		(\bm{v}_{i})}
	{\partial \bm{v}_{i}} 
	= 
	\frac{
		\sum_{j=1, i \neq j}^{B} 
		(\bm{t}_{j} - \bm{t}_{i})
		\cdot
		\mathcal{D}_{\text{Unified}}(\bm{t}_{j})}
	{1 + \sum_{j=1, i \neq j}^{B} 
		\mathcal{D}_{\text{Unified}}(\bm{t}_{j})},
\end{equation}
\begin{equation}
	\frac{
		\partial 
		\mathcal{L}_{\text{Unified}}
		(\bm{v}_{i})}
	{\partial \bm{t}_{j}} 
	= \bm{v}_{i}
	\cdot
	\frac{\mathcal{D}_{\text{Unified}}(\bm{t}_{j})}
	{1 + \mathcal{D}_{\text{Unified}}(\bm{t}_{j})},
\end{equation}
where
\begin{equation}
	\mathcal{D}_{\text{Unified}}(\bm{t}_{j})
	= e^{  
		\left( \bm{v}_{i}^{\top} \bm{t}_{j}
		- \bm{v}_{i}^{\top} \bm{t}_{i} + m
		\right)}.
\end{equation}
The modulo value of the component of the gradient in the direction of the hypersphere tangent can be expressed as:
\begin{equation}
	\begin{aligned}
		\left \Vert \bm{g}_{\text{Unified}}(\bm{v}_{i}) 
		\right \Vert 
		=
		& \sum_{j=1, i \neq j}^{B} 
		\left \Vert
		\bm{t}_{j} - \bm{t}_{i}
		\right \Vert 
		\cdot \sin 
		\theta(\bm{v}_{i}, \bm{t}_{j} - \bm{t}_{i}) \\
		& \cdot \frac{ \mathcal{D}_{\text{Unified}}(\bm{t}_{j}) }
		{1 + \sum_{j=1, i \neq j}^{B} 
			\mathcal{D}_{\text{Unified}}(\bm{t}_{j})},
	\end{aligned}
\end{equation}
\begin{equation}
	\left \Vert
	\bm{g}_{\text{Unified}}(\bm{t}_{j}) \right \Vert 
	=
	\left \Vert
	\frac{
		\partial 
		\mathcal{L}_{\text{Unified}}
		(\bm{v}_{i})}
	{\partial \bm{t}_{j}} 
	\right \Vert 
	\cdot \sin 
	\theta(\bm{v}_{i}, \bm{t}_{j}).
\end{equation}
Since our unified loss considers all negative samples, $ \exists \bm{t}_{j} $, $ \bm{t}_{i}^{\top} \bm{t}_{j} \nrightarrow 1 $, $ \Vert \bm{g}_{\text{Unified}}(\bm{v}_{i})  \Vert \nrightarrow 0 $.
Besides, $ \exists \bm{t}_{j} $, $ \sin 
\theta(\bm{v}_{i}, \bm{t}_{j}) \nrightarrow 0 $, $ \Vert \bm{g}_{\text{Unified}}(\bm{t}_{j}) \Vert \nrightarrow 0 $.
Therefore, the gradient of the model trained with unified loss does not vanish, and the model converges faster.

The gradient of VLC is similar to the unified loss, the only difference is:
\begin{equation}
	\mathcal{D}_{\text{VLC}}(\bm{t}_{j}) 
	= e^{  
	\left( \bm{v}_{i}^{\top} \bm{t}_{j}
	- \bm{v}_{i}^{\top} \bm{t}_{i}
	\right)}.
\end{equation}
Therefore, under the same conditions, $ \mathcal{L}_{\text{Unified}} $ can contribute a larger gradient than $ \mathcal{L}_{\text{VLC}} $, making positive and negative samples more separated and having better generalization. 

\begin{table*}[t]
	\caption{Experimental results (\%) on Flickr30K and MS-COCO 1K. *: Ensemble results of two models.}
	\setlength\tabcolsep{3pt}
	\small
	\begin{center}
		\begin{tabular}{cccccccccccccccc}
			\toprule[1pt]
			Data Split
			& \multicolumn{7}{c}{Flickr30K 1K Test} & & \multicolumn{7}{c}{MS-COCO 5-fold 1K Test} \\
			\cline{2-8}\cline{10-16}
			%			\specialrule{0em}{2pt}{0pt}
			Eval Task 
			& \multicolumn{3}{c}{Image-to-Text} & \multicolumn{3}{c}{Text-to-Image} & \multirow{2}*{RSUM} & & \multicolumn{3}{c}{Image-to-Text} & \multicolumn{3}{c}{Text-to-Image} & \multirow{2}*{RSUM} \\
			\cline{2-7}\cline{10-15}
			\specialrule{0em}{2pt}{0pt}
			Method 
			& R@1 & R@5 & R@10 & R@1 & R@5 & R@10 & & & R@1 & R@5 & R@10 & R@1 & R@5 & R@10 \\
			\hline
			
			\specialrule{0em}{2pt}{0pt}
			SCAN* \cite{lee2018stacked} & 67.4 & 90.3 & 95.8 & 48.6 & 77.7 & 85.2 & 465.0 & & 72.7 & 94.8 & 98.4 & 58.8 & 88.4 & 94.8 & 507.9 \\
			VSRN* \cite{li2019visual} & 71.3 & 90.6 & 96.0 & 54.7 & 81.8 & 88.2 & 482.6 & & 76.2 & 94.8 & 98.2 & 62.8 & 89.7 & 95.1 & 516.8 \\
			CVSE \cite{wang2020consensus} & 73.5 & 92.1 & 95.8 & 52.9 & 80.4 & 87.8 & 482.5 & & 74.8 & 95.1 & 98.3 & 59.9 & 89.4 & 95.2 & 512.7 \\
			CAAN \cite{zhang2020context} & 70.1 & 91.6 & 97.2 & 52.8 & 79.0 & 87.9 & 478.6 & & 75.5 & 95.4 & 98.5 & 61.3 & 89.7 & 95.2 & 515.6 \\
			IMRAM* \cite{chen2020imram} &  74.1 & 93.0 & 96.6 & 53.9 & 79.4 & 87.2 & 484.2 & & 76.7 & 95.6 & 98.5 & 61.7 & 89.1 & 95.0 & 516.6 \\
			GSMN* \cite{liu2020graph} & 76.4 & 94.3 & 97.3 & 57.4 & 82.3 & 89.0 & 496.8 & & 78.4 & 96.4 & 98.6 & 63.3 & 90.1 & 95.7 & 522.5 \\
			VSE$\infty$ \cite{chen2021learning} & 76.5 & 94.2 & 97.7 & 56.4 & 83.4 & 89.9 & 498.1 & & 78.5 & 96.0 & 98.7 & 61.7 & 90.3 & 95.6 & 520.8 \\
			
			\hline
						\specialrule{0em}{2pt}{0pt}
			VSE++ (BUTD) & 69.4 & 90.7 & 95.4 & 52.1 & 79.0 & 85.5 & 472.1 & & 73.0 & 94.5 & 98.2 & 58.3 & 88.1 & \textbf{94.4} & 506.6 \\
			\rowcolor{black!10}
			\textbf{VSE++ (BUTD) + $\mathcal{L}_{\text{Unified}}$} & \textbf{70.7} & \textbf{90.8} & \textbf{95.6} & \textbf{52.9} & \textbf{79.8} & \textbf{86.7} & \textbf{476.4} & & \textbf{74.0} & \textbf{94.6} & \textbf{98.4} & \textbf{58.7} & \textbf{88.6} & \textbf{94.4} & \textbf{508.6} \\
			
			\hline
			%			\specialrule{0em}{2pt}{0pt}
			BFAN* \cite{liu2019focus} & 68.1 & 91.4 & 95.9 & 50.8 & 78.4 & 85.8 & 470.4 & & 74.9 & 95.2 & 98.3 & 59.4 & 88.4 & 94.5 & 510.7 \\
			\rowcolor{black!10}
			\textbf{BFAN* + $\mathcal{L}_{\text{Unified}}$} & \textbf{74.3} & \textbf{93.8} & \textbf{96.7} & \textbf{54.5} & \textbf{80.8} & \textbf{87.5} & \textbf{487.6} & & \textbf{76.2} & \textbf{95.8} & \textbf{98.7} & \textbf{60.7} & \textbf{88.6} & \textbf{94.7} & \textbf{514.7} \\
			
			\hline
			%			\specialrule{0em}{2pt}{0pt}
			SGRAF* \cite{diao2021similarity} & 77.8 & 94.1 & \color{blue}\textbf{97.4} & 58.5 & 83.0 & 88.8 & 499.6 & & 79.6 & 96.2 & 98.5 & 63.2 & 90.7 & \color{blue}\textbf{96.1} & 524.3 \\
			\rowcolor{black!10}
			\textbf{SGRAF* + $\mathcal{L}_{\text{Unified}}$} & \color{blue}\textbf{78.3} & \color{blue}\textbf{95.0} & \color{blue}\textbf{97.4} & \color{blue}\textbf{60.4} & \color{blue}\textbf{85.0} & \color{blue}\textbf{90.6} & \color{blue}\textbf{506.6} & & \color{blue}\textbf{79.9} & \color{blue}\textbf{97.0} & \color{blue}\textbf{98.8} & \color{blue}\textbf{65.1} & \color{blue}\textbf{90.8} & 96.0 & \color{blue}\textbf{527.4} \\
			
			\bottomrule[1pt]
		\end{tabular}
	\end{center}
	\label{itr}
\end{table*}
\begin{table*}[t]
	\caption{Experimental results (\%) on Flickr30K and MS-COCO 5K.}
	\setlength\tabcolsep{3pt}
	\small
	\begin{center}
		\begin{tabular}{cccccccccccccccc}
			\toprule[1pt]
			Data Split
			& \multicolumn{7}{c}{Flickr30K 1K Test} & & \multicolumn{7}{c}{MS-COCO 5K Test} \\
			\cline{2-8}\cline{10-16}
			%			\specialrule{0em}{2pt}{0pt}
			Eval Task 
			& \multicolumn{3}{c}{Image-to-Text} & \multicolumn{3}{c}{Text-to-Image} & \multirow{2}*{RSUM} & & \multicolumn{3}{c}{Image-to-Text} & \multicolumn{3}{c}{Text-to-Image} & \multirow{2}*{RSUM} \\
			\cline{2-7}\cline{10-15}
			%			\specialrule{0em}{2pt}{0pt}
			Method 
			& R@1 & R@5 & R@10 & R@1 & R@5 & R@10 & & & R@1 & R@5 & R@10 & R@1 & R@5 & R@10 \\
			
			\hline
			%			\specialrule{0em}{2pt}{0pt}
			CLIP \cite{radford2021learning} & 88.0 & 98.7 & 99.4 & 68.7 & 90.6 & 95.2 & 540.6 & & 58.4 & 81.5 & 88.1 & 37.8 & 62.4 & 72.2 & 400.4 \\
			CLIP + $ \mathcal{L}_{\text{VLC}} $ & 86.7 & 97.6 & 99.6 & 73.3 & 93.6 & 96.6 & 547.4 & & 61.5 & 85.0 & 91.6 & 46.0 & 73.2 & 82.2 & 439.5 \\
			\rowcolor{black!10}
			\textbf{CLIP + $\mathcal{L}_{\text{Unified}}$} & \textbf{89.1} & \textbf{98.6} & \textbf{99.8} & \textbf{76.4} & \textbf{94.3} & \textbf{97.1} & \textbf{555.2} & & \textbf{62.2} & \textbf{85.4} & \textbf{91.9} & \textbf{47.0} & \textbf{74.0} & \textbf{83.0} & \textbf{443.5} \\
			
			\bottomrule[1pt]
		\end{tabular}
	\end{center}
	\label{vlp}
\end{table*}
\section{Experiments}
\subsection{Dataset and Experiment Settings}
Vision-language retrieval mainly includes image-text and video-text retrieval, and we conduct experiments on these two tasks.
For image-text retrieval, we evaluate our loss on two benchmarks: Flickr30K \cite{young2014image} and MS-COCO \cite{lin2014microsoft}.
Flickr30K contains 31,000 images, each image is annotated with 5 sentences.
We use 1,000 images for validation, 1,000 images for testing and the remaining 29,000 images for training. 
MS-COCO contains 123,287 images, each image associates with 5 sentences. 
We use 113,287 images for training, 5,000 images for validation and 5,000 images for testing. 
We report results on both 1,000 test images (averaged over 5 folds) and full 5,000 test images of MS-COCO. 
For video-text retrieval, we evaluate our loss on MSR-VTT \cite{xu2016msr},
which includes 10,000 videos, each video comes with 20 sentence descriptions. 
We evaluate the performance on two splits, \textit{i.e.}, the ``full'' and ``1k-A'' splits.
The ``full'' split uses 6,513 videos for training, 497 videos for validation, and 2,990 videos for testing. 
The other ``1k-A'' split \cite{yu2018joint} uses 9,000 videos for training, and the remaining 1,000 videos are used for testing.
More implementation details are listed in the Appendix.

\subsection{Image-Text Retrieval}
\subsubsection{Image-Text Retrieval Without Pre-training}
In order to justify the superiority of our unified loss over the state-of-the-art image-text retrieval models, we conduct experiments on VSE++, BFAN \cite{liu2019focus} and SGRAF \cite{diao2021similarity} by only replacing the loss functions.
\begin{itemize}
	\item \textbf{VSE++} is the most representative model in image-text retrieval and several variant models are proposed based on this method.
	We re-implement VSE++ using bottom-up and top-down attention region features \cite{anderson2018bottom} as our baseline, denoted as VSE++ (BUTD).
	\item \textbf{BFAN} is a bidirectional focal attention network for image-text retrieval that aligns image and text by focusing on relevant segments.
	\item 
	\textbf{SGRAF} is a similarity graph reasoning and attention filtration network for image-text retrieval. 
	SGRAF is one of the open-sourced state-of-the-art models.
\end{itemize}
\tablename~\ref{itr} compares our unified loss with image-text retrieval baselines.
By replacing the loss function with our unified loss, the performance of the three baseline methods is improved.
On Flickr30K dataset, \textbf{SGRAF* + $\mathcal{L}_{\text{Unified}}$} improves RSUM by 7.0\% compared to SGRAF*. 
On MS-COCO 1K test set, applying our unified loss to SGRAF* can improve RSUM by 3.1\%.
%Due to the limitation of the main text, please refer to the appendix for the experimental comparison results of more datasets and methods.
Please refer to the Appendix for the experimental comparison results of more datasets and methods.

\subsubsection{Fine-tuning on Vision-Language Pre-training Model}
In order to verify the improvement of our unified loss on the vision-language pre-training model, we also conduct experiments on CLIP \cite{radford2021learning}.
We use the official ViT-L/14@336px model for experiments.
\tablename~\ref{vlp} summarizes the retrieval results on Flickr30K and MS-COCO 5K test set.
CLIP + $ \mathcal{L}_{\text{VLC}} $ denotes  our experimental result of fine-tuning using VLC as the optimization objective.
\textbf{CLIP + $\mathcal{L}_{\text{Unified}}$} denotes the experimental result of fine-tuning using our unified loss as the optimization objective.
Our unified loss yields a 7.8\% increase on Flickr30K and 4.0\% improvement on MS-COCO 5K test sets in terms of RSUM when compared with CLIP + $ \mathcal{L}_{\text{VLC}} $.
This shows that our unified loss is more discriminative and can provide better generalization performance in downstream fine-tuning tasks.

\begin{table}[t]
	\caption{Experimental results (\%) on Flickr30K.}
	\small
	\begin{center}
		\begin{tabular}{ccccccccccccc}
			\toprule[1pt]
			\multirow{2}*{Method}
			& \multicolumn{2}{c}{Image-to-Text} & \multicolumn{2}{c}{Text-to-Image} \\
			\cline{2-5}
%			\specialrule{0em}{2pt}{0pt}
			~ & R@1 & R@5 & R@1 & R@5  \\
			
			\hline
%			\specialrule{0em}{2pt}{0pt}
			SCAN T2I AVG & 61.8 & 87.5 & 45.8 & 74.4 \\
			SCAN T2I AVG + RSP & 66.5 & 91.1 & 51.0 & 76.8 \\
			\rowcolor{black!10}
			\textbf{SCAN T2I + $\mathcal{L}_{\text{Unified}}$} & \textbf{70.7} & \textbf{91.4} & \textbf{52.8} & \textbf{78.0} \\ 
			
			\hline
%			\specialrule{0em}{2pt}{0pt}
			BFAN* & 68.1 & 91.4 & 50.8 & 78.4 \\
			BFAN* + SSP & 71.3 & 92.6 & 52.5 & 79.5 \\
			BFAN* + AOQ & 73.2 & \textbf{94.5} & 54.0 & 80.3 \\
			BFAN* + Meta-SPN & 72.5 & 93.2 & 53.3 & 80.2 \\
			\rowcolor{black!10}
			\textbf{BFAN* + $\mathcal{L}_{\text{Unified}}$} & \textbf{74.3} &  93.8 & \textbf{54.5} & \textbf{80.8}  \\
			
			\hline
%			\specialrule{0em}{2pt}{0pt}
			SGRAF* & 77.8 & 94.1 & 58.5 & 83.0 \\
			SGRAF* + NCR & 77.3 & 94.0 & 59.6 & 84.4 \\
			\rowcolor{black!10}
			\textbf{SGRAF* + $\mathcal{L}_{\text{Unified}}$} & \textbf{78.3} & \textbf{95.0} & \textbf{60.4} & \textbf{85.0} \\
			
			\bottomrule[1pt]
		\end{tabular}
	\end{center}
	\label{other}
\end{table}
\subsubsection{Comparison to the Other Loss Functions for Vision-Language Retrieval}
There are a number of loss functions proposed for vision-language retrieval, so we compare our unified loss with these losses:
\begin{itemize}
	\item \textbf{SSP}: Self-similarity polynomial loss \cite{wei2020universal} is a weighted triplet loss that defines a weight function for the positive and negative pairs respectively.
	\item \textbf{RSP}: Relative-similarity polynomial loss \cite{wei2021universal} is a weighted triplet loss that assigns weights to the relative-similarity scores between sample pairs.
	\item \textbf{Meta-SPN}: Meta-SPN \cite{wei2021meta} is a Meta Self-Paced Network that automatically learns a weighting scheme from data for cross-modal matching.
	\item \textbf{AOQ}: Adaptive offline quintuplet loss \cite{chen2020adaptive} provides offline negatives by sampling negatives offline from the whole training set.
	\item \textbf{NCR}: Noisy correspondence rectifier \cite{huang2021learning} is a method for learning with noisy correspondence for cross-modal matching.
\end{itemize}
For a fair comparison, we use the same model as in the papers of various loss functions above but simply replace the loss function with our unified loss.
The experimental results of unified loss compared to the other losses on Flickr30K are shown in \tablename~\ref{other}.
Compared with other loss functions, unified loss improves most evaluation metrics.
Unified loss does not need to introduce many hyperparameters to assign weights like SSP and RSP, nor does it need to train an additional network for weight assignment like Meta-SPN.
Compared with AOQ, Unified loss only requires online hard negative mining, and does not increase the complexity.
Please refer to the Appendix for the experimental comparison results of more methods.

\begin{table}[t]
	\caption{Experimental results (\%) on MSR-VTT.}
	\setlength\tabcolsep{2.4pt}
	\small
	\begin{center}
		\begin{tabular}{ccccccccccccc}
			\toprule[1pt]
			\multirow{2}*{Method} 
			& \multirow{2}*{Split}
			& \multicolumn{3}{c}{Video-to-Text} & \multicolumn{3}{c}{Text-to-Video} \\
			\cline{3-8}
%			\specialrule{0em}{2pt}{0pt}
			~ & ~ & R@1 & R@5 & R@10 & R@1 & R@5 & R@10  \\
			\hline
			
%			\specialrule{0em}{2pt}{0pt}
			CE & Full & 15.6 & 40.9 & 55.2 & 10.0 & 29.0 & 41.2 \\
			\rowcolor{black!10}
			\textbf{CE + $\mathcal{L}_{\text{Unified}}$} & \textbf{Full} & \textbf{17.4} & \textbf{45.1} & \textbf{58.4} & \textbf{11.7} & \textbf{32.3} & \textbf{44.2} \\
			
			\hline
%			\specialrule{0em}{2pt}{0pt}
			MMT & 1k-A & 24.6 & 54.0 & 67.1 & 24.4 & 56.0 & 67.8 \\
			\rowcolor{black!10}
			\textbf{MMT + $\mathcal{L}_{\text{Unified}}$} & \textbf{1k-A} & \textbf{27.3} & \textbf{57.5} & \textbf{69.6} & \textbf{28.6} & \textbf{57.1} & \textbf{69.5} \\
			
			\bottomrule[1pt]
		\end{tabular}
	\end{center}
	\label{vtr}
\end{table}
\begin{figure}[t]
	\centering
	\subfigure[Loss]{
		\label{train_loss}
		\includegraphics[width=0.46\linewidth]{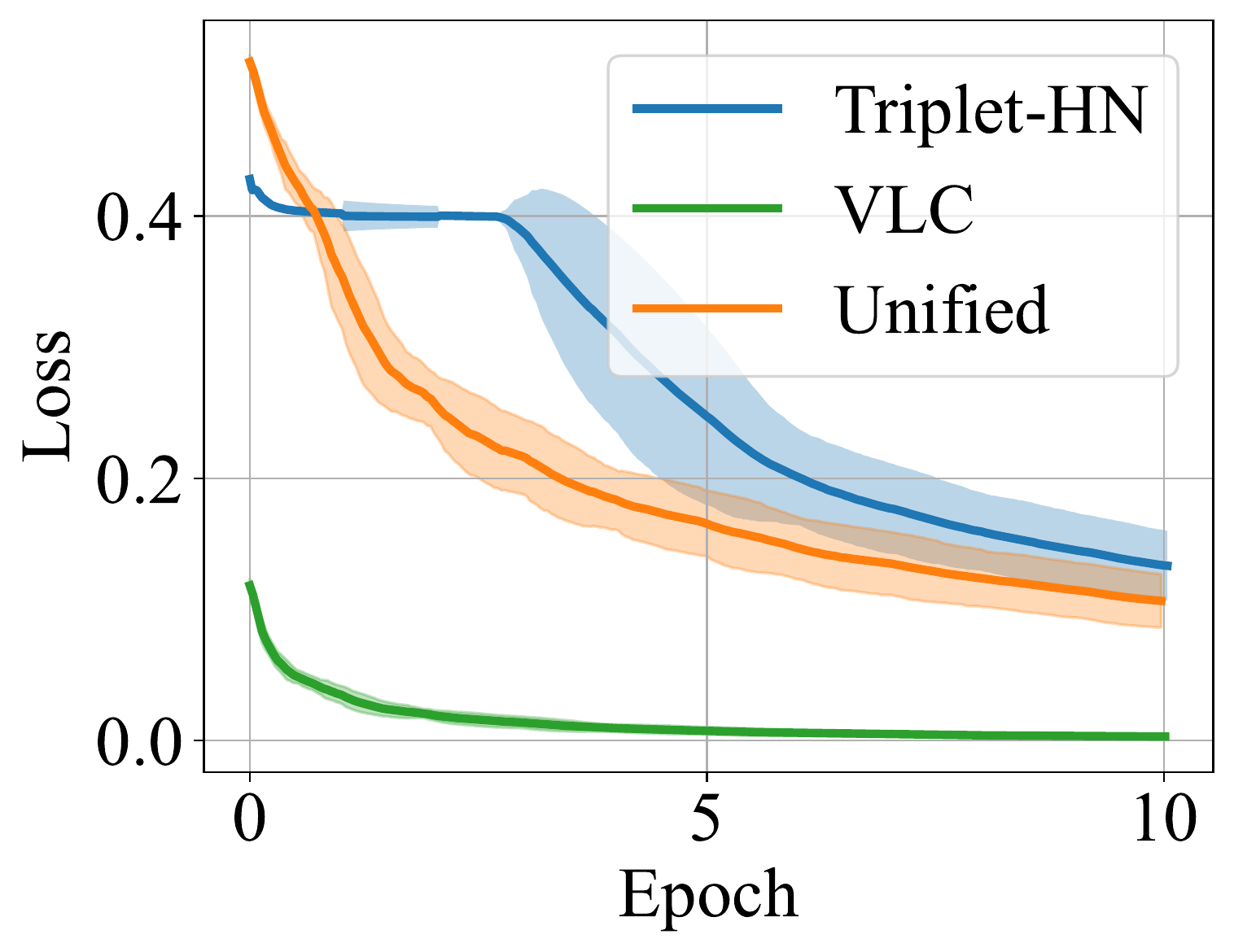}
	}
	\subfigure[RSUM]{
		\label{train_rsum}
		\includegraphics[width=0.46\linewidth]{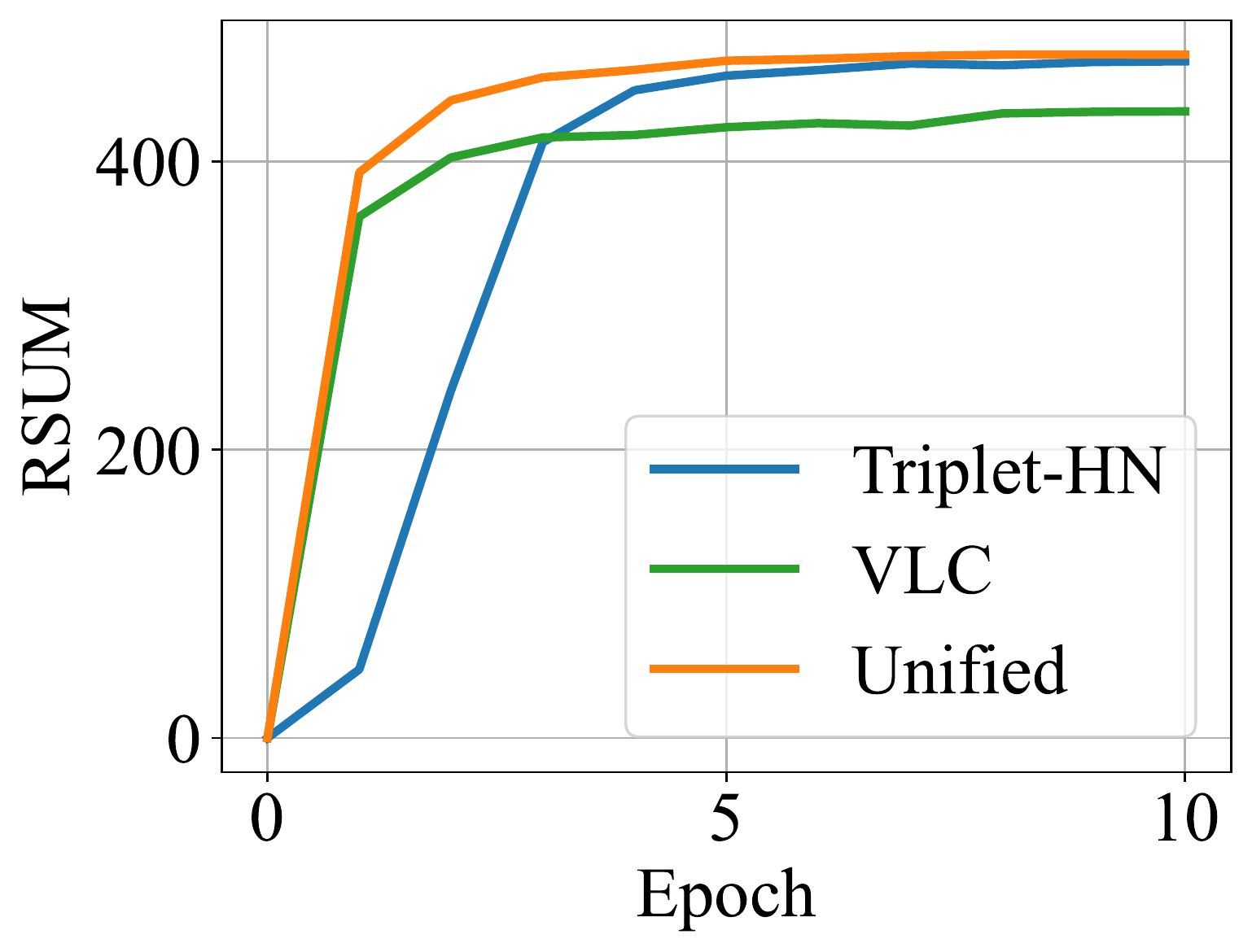}
	}
	\caption{Plotting the loss and RSUM on Flickr30K validation set using VSE++ (BUTD) against epochs.}
\end{figure}
\begin{figure*}[t]
	\centering
	\subfigure[Triplet-HN]{
		\label{train_triplet_sim}
		\includegraphics[width=0.23\linewidth]{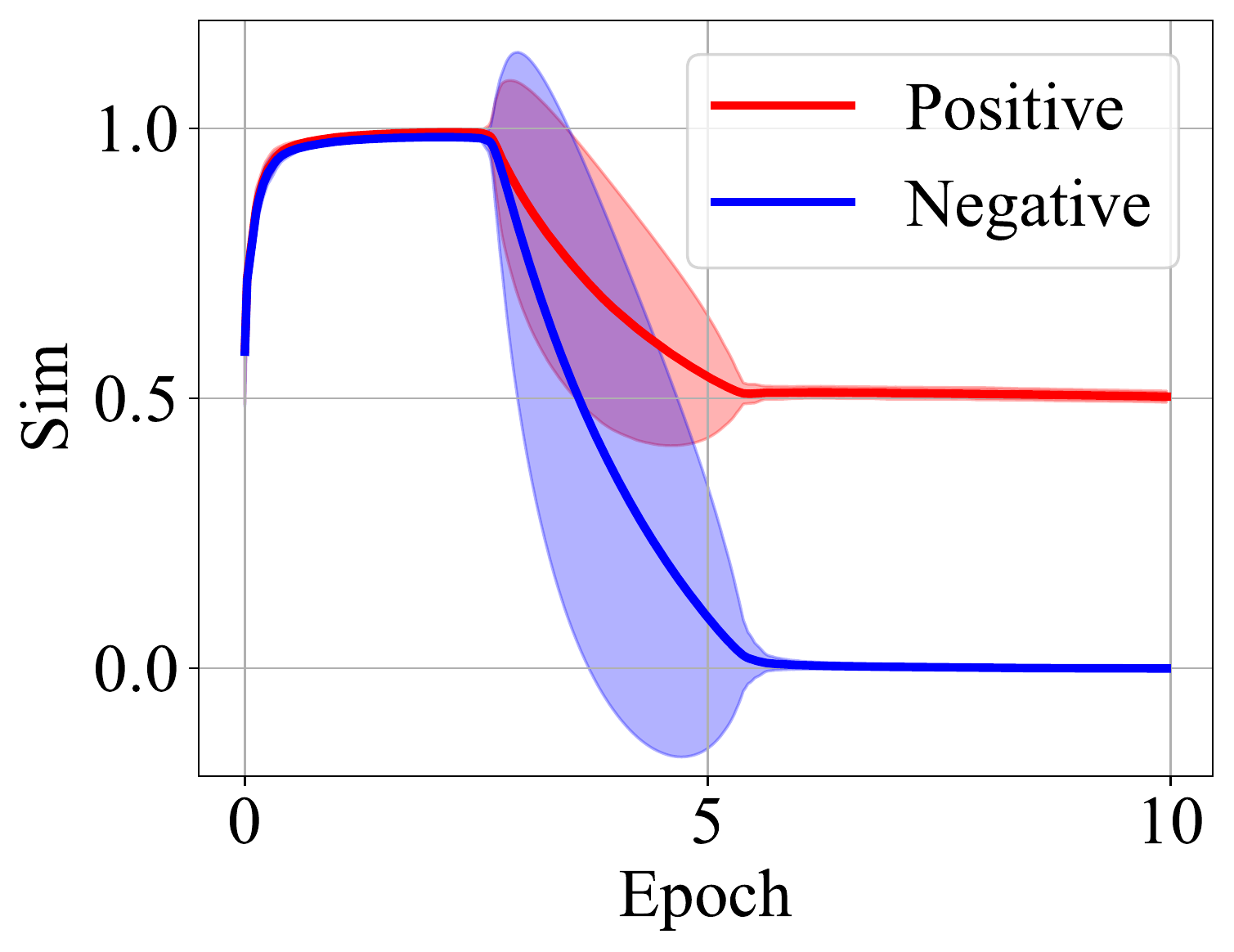}
	}
	\subfigure[VLC]{
		\label{train_vlc_sim}
		\includegraphics[width=0.23\linewidth]{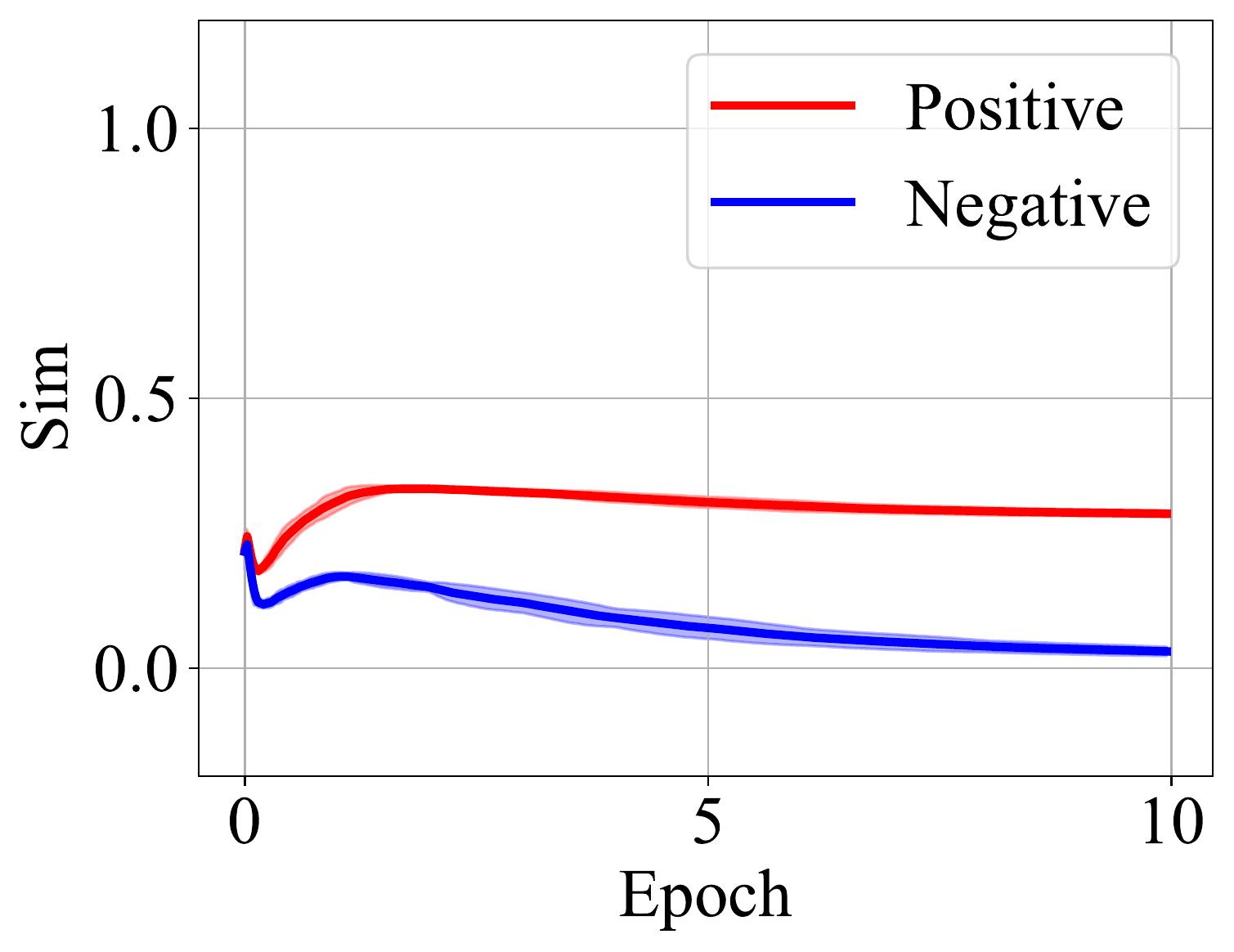}
	}
	\subfigure[Unified]{
		\label{train_unified_sim}
		\includegraphics[width=0.23\linewidth]{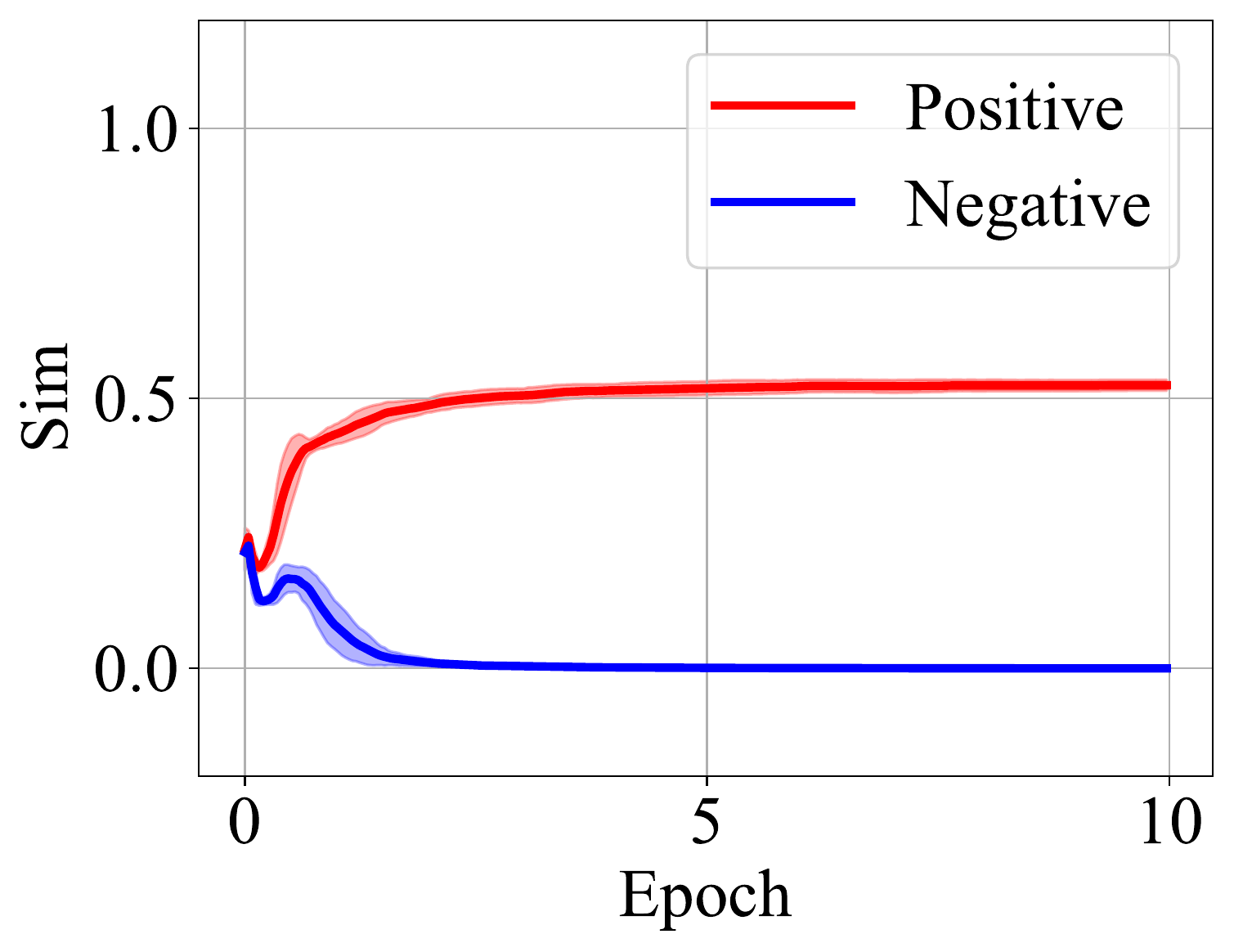}
	}
	\caption{Plotting training epoch against similarities of positive and negative sample pairs on Flickr30K validation set.}
\end{figure*}
\subsection{Video-Text Retrieval}
In order to verify the improvement of our unified loss on the state-of-the-art video-text retrieval models, we conduct experiments on MMT \cite{gabeur2020multi} and CE \cite{liu2019use} by only replacing the loss functions.
\begin{itemize}
	\item MMT is a multi-modal transformer that  jointly encodes the different modalities in video. 
	\item CE is a collaborative experts model that  aggregates information from these different pre-trained experts.
\end{itemize}
\tablename~\ref{vtr} compares our unified loss with video-text retrieval baselines on MSR-VTT.
After replacing the loss with our unified loss, the performance of both models is greatly improved.
The RSUM of MMT using unified loss on 1k-A split is improved by 15.7 \%.
The RSUM of \textbf{CE + $\mathcal{L}_{\text{Unified}}$} on full split is improved by 17.2 \%.
Extensive experiments on multiple image-text and video-text retrieval models show that our unified loss can significantly improve the performance of the state-of-the-art vision-language retrieval models.

\subsection{Convergence Analysis}
\figurename~\ref{train_loss} and \figurename~\ref{train_rsum} compare the performance of Triplet-HN, VLC and unified loss during training.
We plot the loss and RSUM on Flickr30K validation set using VSE++ (BUTD) against epochs.
It can be seen from \figurename~\ref{train_loss} that our unified loss and VLC has better convergence than Triplet-HN. 
\figurename~\ref{train_rsum} shows that the retrieval model can achieve superior retrieval performance faster using unified loss.
In order to further explore the difference in the convergence of different loss functions and compare their generalization, we plot similarities of positive and hardest negative sample pairs on Flickr30K validation set  against epochs.
\figurename~\ref{train_triplet_sim} shows the performance of Triplet-HN during training.
At the beginning of training (epoch $ 0 \sim 3 $), the similarities between the positive pairs and the negative pairs both approaches $ 1 $. 
In this case, the gradients for training approach $ 0 $, resulting in vanishing gradient.
This is the reason why Triplet-HN is easy to fall into local minima and affect the convergence of training.
VLC and unified loss do not have the above problems since they consider all negative samples to participate in training, and show better convergence.
By comparing \figurename~\ref{train_vlc_sim} and \figurename~\ref{train_unified_sim}, it can be seen that unified loss can make the difference in similarity between positive and negative samples larger, thanks to the margin $ m $ introduced for better similarity separation.
Compared with the VLC, our unified loss is more discriminative and can provide better generalization.

\subsection{Parameter Analysis}
\begin{figure}[t]
	\centering
	\subfigure[Margin $ m $]{
		\label{margin_rsum}
		\includegraphics[width=0.46\linewidth]{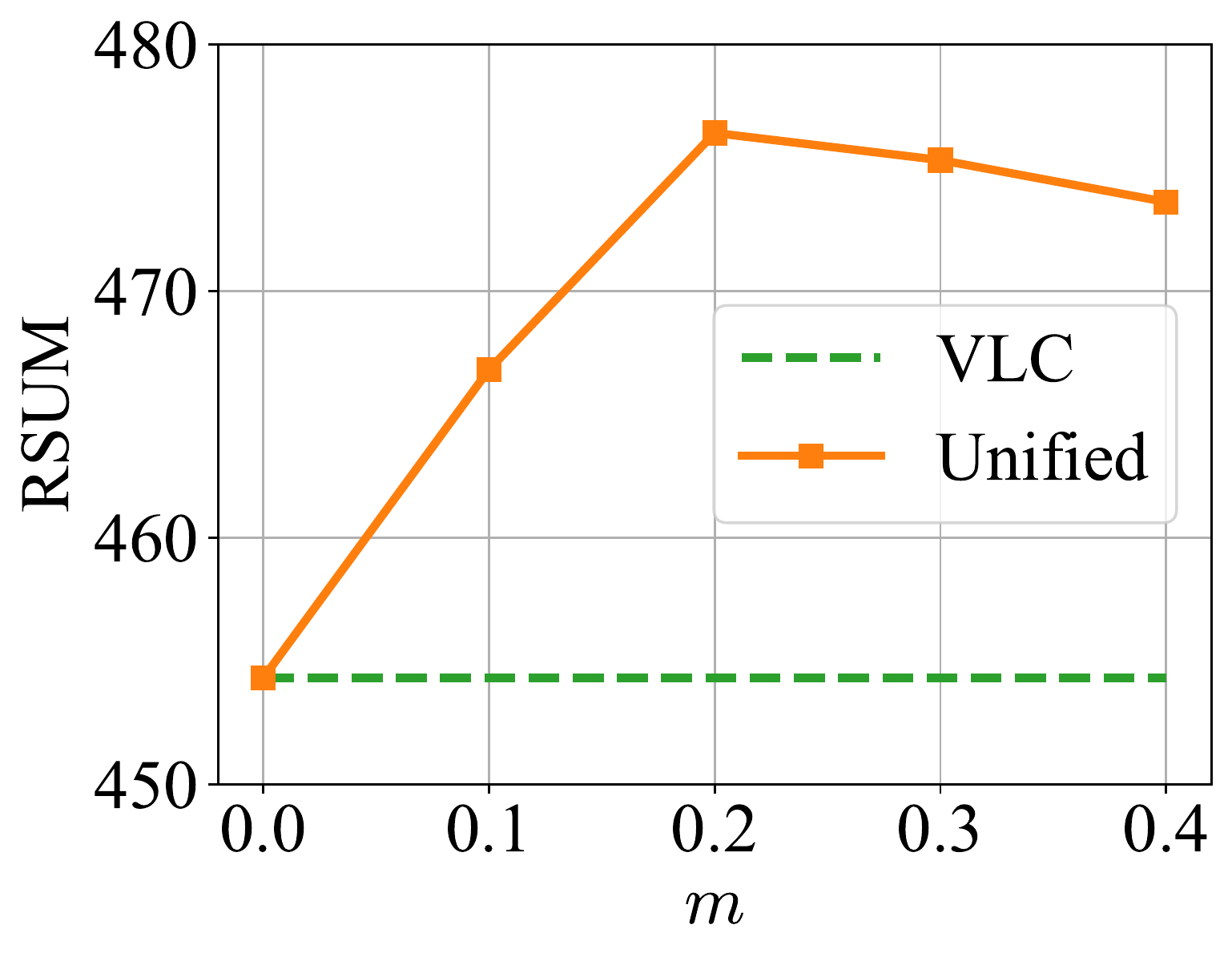}
	}
	\subfigure[Scale factor $ \gamma $]{
		\label{gamma_rsum}
		\includegraphics[width=0.46\linewidth]{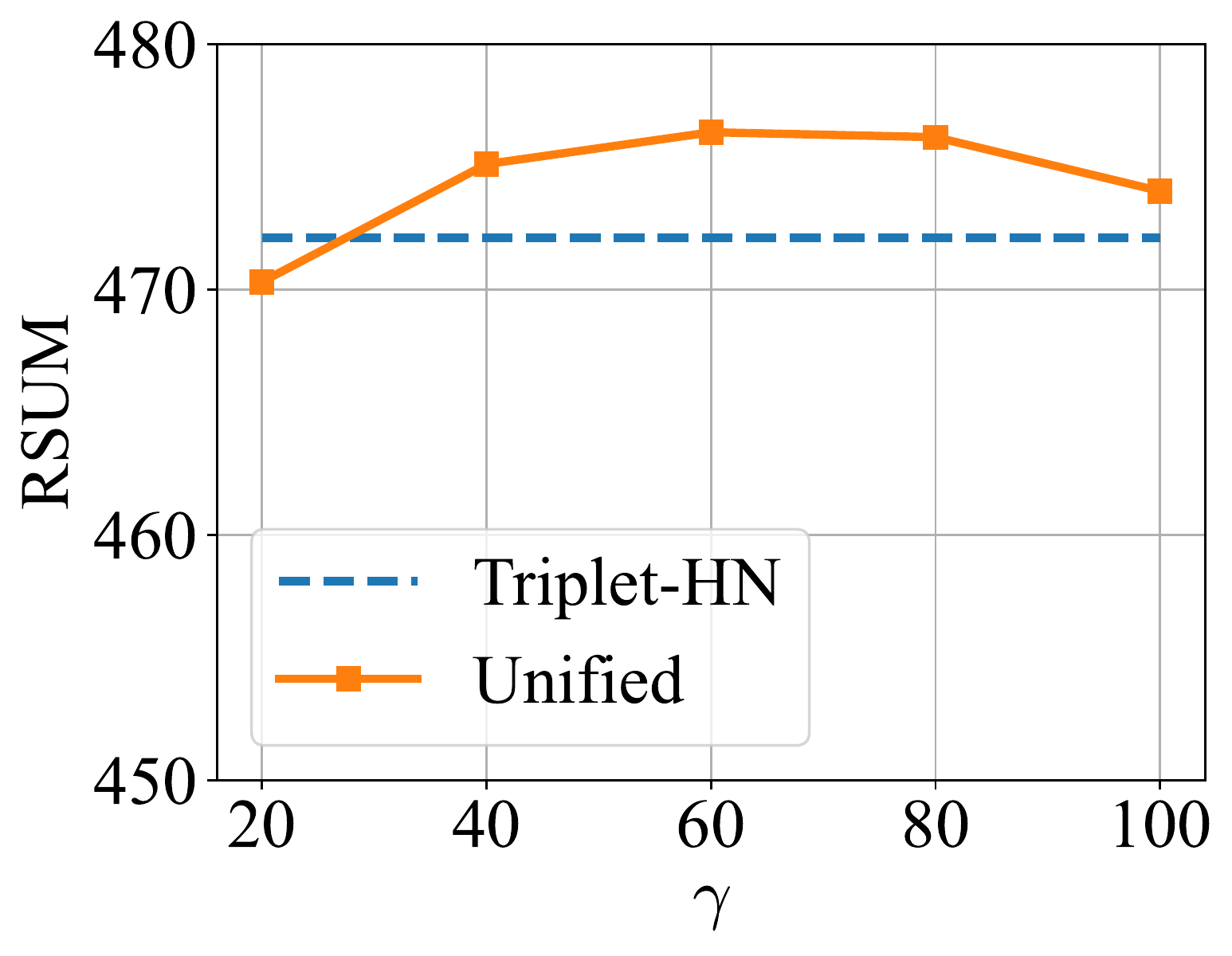}
	}
	\caption{Effects of different configurations of hyper-parameters on Flickr30K.}
\end{figure}
There are two parameters, \textit{i.e.}, margin $ m $ and scale factor $ \gamma $, in our unified loss can be tuned. 
We experiment with several combinations of parameters on Flickr30K using VSE++ (BUTD).
We test the effect of $ m $ by fixing $ \gamma = 60 $.
We test RSUM on the test set under different parameter conditions. 
As shown in \figurename~\ref{margin_rsum}, when $ m = 0.2 $, the retrieval model achieves the best performance.
When $ m $ takes a value greater than 0, the retrieval performance of our unified loss is always better than that of VLC.
We test the effect of $ \gamma $ by fixing $ m = 0.2 $. 
From \figurename~\ref{gamma_rsum} we can see that as $ \gamma $ increases, the retrieval performance first increases and then decreases.
Except for the case of $ \gamma=20 $, the retrieval performance of our unified loss is always better than Triplet-HN in other cases.
This shows that mining hard negative samples is essential for improving retrieval performance. 
However, excessive mining of hard negative samples, such as Triplet-HN only mining one hardest negative samples, will also lose performance.

\section{Conclusion}
This paper proposes a unified loss of pair similarity optimization for vision-language retrieval, providing a powerful tool for understanding existing loss functions. 
Most of the existing loss functions are shown to be variants of the proposed unified loss. 
We find that most of the loss functions are essentially minimize the difference between the similarities of negative pairs and the positive pairs.
On the other hand, our unified loss combines the advantages of triplet loss and contrastive learning loss, and overcomes their respective problems.
Compared with existing loss functions, our unified loss has better convergence and generalization ability.
In the future, we will do further work on the hyper-parameter selection of the unified loss.

\cleardoublepage
\appendix
In this appendix, we provide equation derivations, implementation details, and experiments omitted from the main text.

\section{Equation Derivation}
\subsection{Convert VLC into a Form that Optimizes Pair Similarity}
VLC can be converted into a form that optimizes pair similarity:
\begin{equation} \label{vlc_der}
	\begin{aligned}
		\mathcal{L}_{\text{VLC}}
		= & - \sum_{i=1}^{B}
		\left(
		\log 
		\dfrac{e^{\gamma s_{ii}}}
		{\sum_{j=1}^{B} e^{\gamma s_{ij}}}
		+ \log 
		\dfrac{e^{\gamma s_{ii}}}
		{\sum_{j=1}^{B} e^{\gamma s_{ji}}} \right) \\
		= & \sum_{i=1}^{B}
		\left(
		\log 
		\dfrac
		{\sum_{j=1}^{B} e^{\gamma s_{ij}}}
		{e^{\gamma s_{ii}}}
		+ \log 
		\dfrac
		{\sum_{j=1}^{B} e^{\gamma s_{ji}}}
		{e^{\gamma s_{ii}}} 
		\right) \\
		= & \sum_{i=1}^{B}
		\left(
		\log
		\left( 1 + \sum_{j=1, i \neq j}^{B} 
		e^{ 
			\gamma \left( s_{ij} - s_{ii} 
			\right)} \right) \right. \\
		& + \left. \log 
		\left( 1 + \sum_{j=1, i \neq j}^{B} 
		e^{
			\gamma \left( s_{ji} - s_{ii} \right)
		} \right) \right).
	\end{aligned}
\end{equation}
From Eq.~\eqref{vlc_der} we find that $ \mathcal{L}_{\text{VLC}} $ is essentially optimizing the differences $ (s_{ij} - s_{ii}) $ and $ (s_{ji} - s_{ii}) $ between the similarities of negative pairs $ s_{ij} $ and $ s_{ji} $ and positive pair $ s_{ii} $.
The optimization goal of $ \mathcal{L}_{\text{VLC}} $ is to make $ s_{ii} > s_{ij} $ and $ s_{ii} > s_{ji} $, which does not leave enough margin for the distinction between positive and negative samples.
Therefore, $ \mathcal{L}_{\text{VLC}} $ has insufficient discriminative ability for positive and negative samples, which affects the generalization ability of the model, especially the performance of fine-tuning on small datasets.

\newpage
\subsection{Convert Unified Loss into Triplet-HN}
Widely used $ \mathcal{L}_{\text{Triplet-HN}} $ is a special forms of $ \mathcal{L}_{\text{Unified}} $. 
When $ \gamma \rightarrow +\infty $, $ \mathcal{L}_{\text{Unified}} $ is transformed into $ \mathcal{L}_{\text{Triplet-HN}} $:
\begin{equation}
	\begin{aligned}
		\mathcal{L}_{\text{Triplet-HN}}
		= & \lim_{\gamma \rightarrow +\infty}
		\mathcal{L}_{\text{Unified}} \\
		= & \lim_{\gamma \rightarrow +\infty}
		\frac{1}{\gamma} \sum_{i=1}^{B}
		\left(
		\log
		\left( 1 + \sum_{j=1, i \neq j}^{B} 
		e^{ 
			\gamma \left( s_{ij} - s_{ii} + m
			\right)} \right) \right. \\
		& + \left. \log 
		\left( 1 + \sum_{j=1, i \neq j}^{B} 
		e^{
			\gamma \left( s_{ji} - s_{ii} + m \right)
		} \right) \right) \\
		= & \lim_{\gamma \rightarrow +\infty}
		\frac{1}{\gamma} \sum_{i=1}^{B}
		\left(
		\log
		\left( 1 +
		e^{ 
			\gamma \left( s(V_{i}, \hat{T}_{i}) - s(V_{i}, T_{i}) + m 
			\right)} \right) \right. \\
		& + \left. \log 
		\left( 1 +
		e^{
			\gamma \left( s(\hat{V}_{i}, T_{i}) - s(V_{i}, T_{i}) + m  \right)
		} \right) \right) \\
		= & \sum_{i=1}^{B}
		\left(
		\left[ 
		s(V_{i}, \hat{T}_{i}) - s(V_{i}, T_{i}) + m 
		\right]_{+}
		\right. \\
		& \left. 
		+ \left[ 
		s(\hat{V}_{i}, T_{i}) - s(V_{i}, T_{i}) + m 
		\right]_{+}
		\right),
	\end{aligned}
\end{equation}
where 
\begin{equation}
	\begin{aligned}
		\hat{T}_{i} = \max_{j=1, i \neq j}^{B} s{(V_{i}, T_{j})}, \\
		\hat{V}_{i} = \max_{j=1, i \neq j}^{B} s{(V_{j}, T_{i})}.
	\end{aligned}
\end{equation}
$ \gamma \rightarrow +\infty $ represents that the loss only mines the hardest negative sample in a batch.
Our unified loss uses $ \gamma $ to control the hardness of mining samples, introducing more negative samples in training.
Compared with Triplet-HN, the unified loss converges faster.

\newpage
\subsection{Convert Unified Loss into VLC}
VLC is also a special forms of our unified loss. 
When $ m = 0 $, $ \mathcal{L}_{\text{Unified}} $ is transformed into $ \mathcal{L}_{\text{VLC}} $:
\begin{equation}
	\begin{aligned}
		\mathcal{L}_{\text{VLC}}
		= & \gamma \cdot
		\mathcal{L}_{\text{Unified}}(m=0) \\
		= & \sum_{i=1}^{B}
		\left(
		\log
		\left( 1 + \sum_{j=1, i \neq j}^{B} 
		e^{ 
			\gamma \left( s_{ij} - s_{ii}
			\right)} \right) \right. \\
		& + \left. \log 
		\left( 1 + \sum_{j=1, i \neq j}^{B} 
		e^{
			\gamma \left( s_{ji} - s_{ii} \right)
		} \right) \right) \\
		= & \sum_{i=1}^{B}
		\left(
		\log
		\left( 1 + \dfrac
		{\sum_{j=1, i \neq j}^{B} e^{\gamma s_{ij}}}
		{e^{\gamma s_{ii}}} \right) \right. \\
		& + \left. \log 
		\left( 1 + \dfrac
		{\sum_{j=1, i \neq j}^{B} e^{\gamma s_{ji}}}
		{e^{\gamma s_{ii}}} \right) \right) \\
		= & \sum_{i=1}^{B}
		\left(
		\log 
		\dfrac
		{\sum_{j=1}^{B} e^{\gamma s_{ij}}}
		{e^{\gamma s_{ii}}}
		+ \log 
		\dfrac
		{\sum_{j=1}^{B} e^{\gamma s_{ji}}}
		{e^{\gamma s_{ii}}} 
		\right) \\
		= & - \sum_{i=1}^{B}
		\left(
		\log 
		\dfrac{e^{\gamma s_{ii}}}
		{\sum_{j=1}^{B} e^{\gamma s_{ij}}}
		+ \log 
		\dfrac{e^{\gamma s_{ii}}}
		{\sum_{j=1}^{B} e^{\gamma s_{ji}}} \right).
	\end{aligned}
\end{equation}
Our unified loss introduces a margin $ m $ for better similarity separation, which can improve the generalization of the retrieval model. 

\newpage
\subsection{The Conversion Between More Loss Functions and Unified Loss}
\subsubsection{Weighted Unified Loss}
There is a lot of work \cite{wei2020universal, chen2020adaptive, wei2021universal, wei2021meta} that proposes weighting metric learning frameworks for vision-language retrieval, which can further improve retrieval performance. 
They can be included in our unified loss, but the similarities of the sample pairs are multiplied by the weights.
\begin{equation}
	\begin{aligned}
		\mathcal{L}_{\text{Unified}}^{\text{Weight}}
		= \frac{1}{\gamma} \sum_{i=1}^{B}
		\left(
		\log
		\left( 1 + \sum_{j=1, i \neq j}^{B} 
		e^{\gamma \left( w_{ij} s_{ij} - w_{ii} s_{ii} + m
			\right)} \right) \right. \\
		+ \left. \log 
		\left( 1 + \sum_{j=1, i \neq j}^{B} 
		e^{\gamma \left(  w_{ji} s_{ji} - w_{ii} s_{ii} + m \right)} \right) \right),
	\end{aligned}
\end{equation}
where $ w_{ij} $ and $ w_{ji} $ are the weights of $ s_{ij} $ and $ s_{ji} $, $ w_{ii} $ is the weight of $ s_{ii} $.
With the proposed unified loss, it is easy to find that most of the loss functions are essentially minimizing the differences $ (s_{ij} - s_{ii}) $ and $ (s_{ji} - s_{ii}) $.

Relative-similarity Polynomial loss (RSP) \cite{wei2021universal} is a weighted triplet loss that assigns appropriate weights to the relative-similarity scores between positive and negative pairs.
It can be transformed into a variant of our unified loss:
\begin{equation}
	\begin{aligned}
		\mathcal{L}_{\text{RSP}}
		= & \sum_{i=1}^{B}
		\left(
		\left[ G_{\text{Rel}} \left(
		\hat{s}_{ij} - s_{ii} \right) 
		+ m \right]_{+} 
		+ \left[ G_{\text{Rel}} \left( 
		\hat{s}_{ji} -  s_{ii} \right) 
		+ m \right]_{+}
		\right) \\
		= & \lim_{\gamma \rightarrow +\infty} \frac{1}{\gamma} \sum_{i=1}^{B}
		\left(
		\log
		\left( 1 + \sum_{j=1, i \neq j}^{B} 
		e^{ 
			\gamma \left( G_{\text{Rel}} 
			\left( s_{ij} -  s_{ii} \right) + m
			\right)} \right) \right. \\
		& + \left. \log 
		\left( 1 + \sum_{j=1, i \neq j}^{B} 
		e^{
			\gamma \left( G_{\text{Rel}} 
			\left( s_{ji} -  s_{ii} \right) + m \right)
		} \right) \right),
	\end{aligned}
\end{equation}
where $ G_{\text{Rel}} $ is the weight to the relative-similarity scores, which is generated by a polynomial function.

\newpage
\subsubsection{Unified Loss with Adaptive Margin}
There is some work \cite{huang2021learning, biten2021image} that proposes that the margin can be adjusted to address the noise problem in vision-language datasets.
They can also be included in our unified loss, but the margin is changed to a variable that can be adjusted adaptively.
\begin{equation}
	\begin{aligned}
		\mathcal{L}_{\text{Unified}}^{\text{Margin}}
		= \frac{1}{\gamma} \sum_{i=1}^{B}
		\left(
		\log
		\left( 1 + \sum_{j=1, i \neq j}^{B} 
		e^{\gamma \left( s_{ij} - s_{ii} + m_{i}
			\right)} \right) \right. \\
		+ \left. \log 
		\left( 1 + \sum_{j=1, i \neq j}^{B} 
		e^{\gamma \left( s_{ji} - s_{ii} + m_{i} \right)} \right) \right),
	\end{aligned}
\end{equation}

Noisy Correspondence Rectifier (NCR) \cite{huang2021learning} is a method for learning with noisy correspondence for cross-modal matching.
\begin{equation}
	\begin{aligned}
		\mathcal{L}_{\text{NCR}}
		= & \sum_{i=1}^{B}
		\left(
		\left[ \hat{s}_{ij} - s_{ii} + m_{i} \right]_{+} 
		+ \left[ \hat{s}_{ji} -  s_{ii} + m_{i} \right]_{+}
		\right) \\
		= & \lim_{\gamma \rightarrow +\infty} \frac{1}{\gamma} \sum_{i=1}^{B}
		\left(
		\log
		\left( 1 + \sum_{j=1, i \neq j}^{B} 
		e^{ 
			\gamma \left( 
			s_{ij} - s_{ii} + m_{i}
			\right)} \right) \right. \\
		& + \left. \log 
		\left( 1 + \sum_{j=1, i \neq j}^{B} 
		e^{
			\gamma \left(
			s_{ji} - s_{ii} + m_{i} \right)
		} \right) \right),
	\end{aligned}
\end{equation}
where $ m_{i} $ is the adaptively adjusted margin.

\clearpage
\begin{table*}[t]
	\caption{Experimental results (\%) on Flickr30K and MS-COCO 1K. *: Ensemble results of two models.}
	\setlength\tabcolsep{3pt}
	\small
	\begin{center}
		\begin{tabular}{cccccccccccccccc}
			\toprule[1pt]
			Data Split
			& \multicolumn{7}{c}{Flickr30K 1K Test} & & \multicolumn{7}{c}{MS-COCO 5-fold 1K Test} \\
			\cline{2-8}\cline{10-16}
			\specialrule{0em}{2pt}{0pt}
			Eval Task 
			& \multicolumn{3}{c}{Image-to-Text} & \multicolumn{3}{c}{Text-to-Image} & \multirow{2}*{RSUM} & & \multicolumn{3}{c}{Image-to-Text} & \multicolumn{3}{c}{Text-to-Image} & \multirow{2}*{RSUM} \\
			\cline{2-7}\cline{10-15}
			\specialrule{0em}{2pt}{0pt}
			Method 
			& R@1 & R@5 & R@10 & R@1 & R@5 & R@10 & & & R@1 & R@5 & R@10 & R@1 & R@5 & R@10 \\
			\hline
			
			\specialrule{0em}{2pt}{0pt}
			SCAN* \cite{lee2018stacked} & 67.4 & 90.3 & 95.8 & 48.6 & 77.7 & 85.2 & 465.0 & & 72.7 & 94.8 & 98.4 & 58.8 & 88.4 & 94.8 & 507.9 \\
			CAMP \cite{wang2019camp} & 68.1 & 89.7 & 95.2 & 51.5 & 77.1 & 85.3 & 466.9 & & 72.3 & 94.8 & 98.3 & 58.5 & 87.9 & 95.0 & 506.8\\ 
			VSRN* \cite{li2019visual} & 71.3 & 90.6 & 96.0 & 54.7 & 81.8 & 88.2 & 482.6 & & 76.2 & 94.8 & 98.2 & 62.8 & 89.7 & 95.1 & 516.8 \\
			DP-RNN \cite{chen2020expressing} & 70.2 & 91.6 & 95.8 & 55.5 & 81.3 & 88.2 & 482.6 & & 75.3 & 95.8 & 98.6 & 62.5 & 89.7 & 95.1 & 517.0 \\
			CVSE \cite{wang2020consensus} & 73.5 & 92.1 & 95.8 & 52.9 & 80.4 & 87.8 & 482.5 & & 74.8 & 95.1 & 98.3 & 59.9 & 89.4 & 95.2 & 512.7 \\
			CAAN \cite{zhang2020context} & 70.1 & 91.6 & 97.2 & 52.8 & 79.0 & 87.9 & 478.6 & & 75.5 & 95.4 & 98.5 & 61.3 & 89.7 & 95.2 & 515.6 \\
			IMRAM* \cite{chen2020imram} &  74.1 & 93.0 & 96.6 & 53.9 & 79.4 & 87.2 & 484.2 & & 76.7 & 95.6 & 98.5 & 61.7 & 89.1 & 95.0 & 516.6 \\
			MMCA \cite{wei2020multi} & 74.2 & 92.8 & 96.4 & 54.8 & 81.4 & 87.8 & 487.4 & & 74.8 & 95.6 & 97.7 & 61.6 & 89.8 & 95.2 & 514.7 \\
			GSMN* \cite{liu2020graph} & 76.4 & 94.3 & 97.3 & 57.4 & 82.3 & 89.0 & 496.8 & & 78.4 & 96.4 & 98.6 & 63.3 & 90.1 & 95.7 & 522.5 \\
			VSE$\infty$ \cite{chen2021learning} & 76.5 & 94.2 & 97.7 & 56.4 & 83.4 & 89.9 & 498.1 & & 78.5 & 96.0 & 98.7 & 61.7 & 90.3 & 95.6 & 520.8 \\
			
			\hline
			\specialrule{0em}{2pt}{0pt}
			VSE++ (BUTD) & 69.4 & 90.7 & 95.4 & 52.1 & 79.0 & 85.5 & 472.1 & & 73.0 & 94.5 & 98.2 & 58.3 & 88.1 & \textbf{94.4} & 506.6 \\
			\rowcolor{black!10}
			\textbf{VSE++ (BUTD) + $\mathcal{L}_{\text{Unified}}$} & \textbf{70.7} & \textbf{90.8} & \textbf{95.6} & \textbf{52.9} & \textbf{79.8} & \textbf{86.7} & \textbf{476.4} & & \textbf{74.0} & \textbf{94.6} & \textbf{98.4} & \textbf{58.7} & \textbf{88.6} & \textbf{94.4} & \textbf{508.6} \\
			
			\hline
			\specialrule{0em}{2pt}{0pt}
			BFAN* \cite{liu2019focus} & 68.1 & 91.4 & 95.9 & 50.8 & 78.4 & 85.8 & 470.4 & & 74.9 & 95.2 & 98.3 & 59.4 & 88.4 & 94.5 & 510.7 \\
			\rowcolor{black!10}
			\textbf{BFAN* + $\mathcal{L}_{\text{Unified}}$} & \textbf{74.3} & \textbf{93.8} & \textbf{96.7} & \textbf{54.5} & \textbf{80.8} & \textbf{87.5} & \textbf{487.6} & & \textbf{76.2} & \textbf{95.8} & \textbf{98.7} & \textbf{60.7} & \textbf{88.6} & \textbf{94.7} & \textbf{514.7} \\
			
			\hline
			\specialrule{0em}{2pt}{0pt}
			SGRAF* \cite{diao2021similarity} & 77.8 & 94.1 & \textbf{97.4} & 58.5 & 83.0 & 88.8 & 499.6 & & 79.6 & 96.2 & 98.5 & 63.2 & 90.7 & \textbf{96.1} & 524.3 \\
			\rowcolor{black!10}
			\textbf{SGRAF* + $\mathcal{L}_{\text{Unified}}$} & \textbf{78.3} & \textbf{95.0} & \textbf{97.4} & \textbf{60.4} & \textbf{85.0} & \textbf{90.6} & \textbf{506.6} & & \textbf{79.9} & \textbf{97.0} & \textbf{98.8} & \textbf{65.1} & \textbf{90.8} & 96.0 & \textbf{527.4} \\
			
			\bottomrule[1pt]
		\end{tabular}
	\end{center}
	\label{itr_full}
\end{table*}
\begin{table*}[t] 
	\caption{Experimental results (\%) on MS-COCO 5K. *: Ensemble results of two models}
	\small
	\begin{center}
		\begin{tabular}{ccccccccccccc}
			\toprule[1pt]
			Eval Task 
			& \multicolumn{3}{c}{Image-to-Text} & \multicolumn{3}{c}{Text-to-Image} & \multirow{2}*{RSUM} \\
			\cline{2-7}
			\specialrule{0em}{2pt}{0pt}
			Method & R@1 & R@5 & R@10 & R@1 & R@5 & R@10 \\
			\hline
			
			\specialrule{0em}{2pt}{0pt}
			SCAN* \cite{lee2018stacked} & 50.4 & 82.2 & 90.0 & 38.6 & 69.3 & 80.4 & 410.9 \\
			VSRN* \cite{li2019visual} & 53.0 & 81.1 & 89.4 & 40.5 & 70.6 & 81.1 & 415.7 \\
			IMRAM* \cite{chen2020imram} & 53.7 & 83.2 & 91.0 & 39.7 & 69.1 & 79.8 & 416.5 \\
			VSE$\infty$ \cite{chen2021learning} & 56.6 & 83.6 & 91.4 & 39.3 & 69.9 & 81.1 & 421.9 \\
			
			\hline
			\specialrule{0em}{2pt}{0pt}
			VSE++ (BUTD) & 49.8 & \textbf{79.5} & 88.5 & \textbf{36.5} & 66.2 & 77.9 & 398.4 \\
			\rowcolor{black!10}
			\textbf{VSE++ (BUTD) + $\mathcal{L}_{\text{Unified}}$} & \textbf{50.6} & 79.4 & \textbf{88.6} & 36.1 & \textbf{66.6} & \textbf{78.4} & \textbf{399.8} \\
			
			\hline
			\specialrule{0em}{2pt}{0pt}
			BFAN* \cite{liu2019focus} & 52.9 & \textbf{82.8} & 90.6 & 38.3 & 67.8 & \textbf{79.3} & 411.7 \\
			\rowcolor{black!10}
			\textbf{BFAN* + $\mathcal{L}_{\text{Unified}}$} & \textbf{54.8} & 82.3 & \textbf{90.8} & \textbf{38.9} & \textbf{68.0} & 79.0 & \textbf{413.9} \\
			
			\hline
			\specialrule{0em}{2pt}{0pt}
			SGRAF* \cite{diao2021similarity} & 57.8 & - & 91.6 & 41.9 & - & 81.3 & - \\
			\rowcolor{black!10}
			\textbf{SGRAF* + $\mathcal{L}_{\text{Unified}}$} & \textbf{59.2} & \textbf{85.5} & \textbf{92.7} & \textbf{43.2} & \textbf{72.7} & \textbf{82.5} & \textbf{429.5} \\
			
			\bottomrule[1pt]
		\end{tabular}
	\end{center}
	\label{itr_coco5k}
\end{table*}
\begin{table*}[t]
	\caption{Experimental results (\%) on Flickr30K. *: Ensemble results of two models.}
	\small
	\begin{center}
		\begin{tabular}{ccccccccccccc}
			\toprule[1pt]
			\multirow{2}*{Method} & \multirow{2}*{Reference}
			& \multicolumn{3}{c}{Image-to-Text} & \multicolumn{3}{c}{Text-to-Image} & \multirow{2}*{RSUM} \\
			\cline{3-8}
			\specialrule{0em}{2pt}{0pt}
			~ & ~ & R@1 & R@5 & R@10 & R@1 & R@5 & R@10  \\
			\hline
			
			\specialrule{0em}{2pt}{0pt}
			SCAN I2T AVG + Triplet-HN \cite{lee2018stacked} & \textit{ECCV} 2018 & 67.9 & 89.0 & 94.4 & 43.9 & 74.2 & 82.8 & 452.2 \\
			SCAN I2T AVG + SSP \cite{wei2020universal} & \textit{CVPR} 2020 & \textbf{69.4} & 89.9 & \textbf{95.4} & 47.5 & 75.5 & 83.1 & 460.8 \\
			\rowcolor{black!10}
			\textbf{SCAN I2T AVG} + \textbf{Unified Loss $\mathcal{L}_{\text{Unified}}$} & \textbf{\textit{Ours}} & 67.3 & \textbf{90.6} & 95.0 & \textbf{47.6} & \textbf{76.7} & \textbf{84.9} & \textbf{462.2} \\
			
			\hline
			\specialrule{0em}{2pt}{0pt}
			SCAN T2I AVG + Triplet-HN \cite{lee2018stacked} & \textit{ECCV} 2018 & 61.8 & 87.5 & 93.7 & 45.8 & 74.4 & 83.0 & 446.2 \\
			SCAN T2I AVG + RSP \cite{wei2021universal} & \textit{TPAMI} 2021 & 66.5 & 91.1 & 95.8 & 51.0 & 76.8 & 84.4 & 465.6 \\
			\rowcolor{black!10}
			\textbf{SCAN T2I AVG} + \textbf{Unified Loss $\mathcal{L}_{\text{Unified}}$} & \textbf{\textit{Ours}} & \textbf{70.7} & \textbf{91.4} & \textbf{95.9} & \textbf{52.8} & \textbf{78.0} & \textbf{85.7} & \textbf{474.5} \\ 
			
			\hline
			\specialrule{0em}{2pt}{0pt}
			BFAN (equal) + Triplet-HN \cite{liu2019focus} & \textit{ACM MM} 2019 & 64.5 & 89.7 & - & 48.8 & 77.3 & - & - \\
			BFAN (equal) + Meta-SPN \cite{wei2021meta} & \textit{ACM MM} 2021 & 69.5 & 92.2 & 96.2 & 50.8 & 78.0 & 85.0 & 471.7 \\
			\rowcolor{black!10}
			\textbf{BFAN (equal)} + \textbf{Unified Loss $\mathcal{L}_{\text{Unified}}$} & \textbf{\textit{Ours}} & \textbf{69.8} & \textbf{93.2} & \textbf{96.7} & \textbf{51.6} & \textbf{78.4} & \textbf{85.6} & \textbf{475.2} \\
			
			\hline
			\specialrule{0em}{2pt}{0pt}
			BFAN (prob) + Triplet-HN \cite{liu2019focus} & \textit{ACM MM} 2019 & 65.5 & 89.4 & - & 47.9 & 77.6 & - & - \\
			BFAN (prob) + Meta-SPN \cite{wei2021meta} & \textit{ACM MM} 2021 & \textbf{68.7} & 90.7 & \textbf{95.8} & 50.0 & 77.1 & 84.7 & 467.0 \\
			\rowcolor{black!10}
			\textbf{BFAN (prob)} + \textbf{Unified Loss $\mathcal{L}_{\text{Unified}}$} & \textbf{\textit{Ours}} & 68.5 & \textbf{92.7} & 95.5 & \textbf{50.8} & \textbf{78.0} & \textbf{85.3}  & \textbf{470.9} \\
			
			\hline
			\specialrule{0em}{2pt}{0pt}
			BFAN* + Triplet-HN \cite{liu2019focus} & \textit{ACM MM} 2019 & 68.1 & 91.4 & 95.9 & 50.8 & 78.4 & 85.8 & 470.4 \\
			BFAN* + SSP \cite{wei2020universal} & \textit{CVPR} 2020 & 71.3 & 92.6 & 96.2 & 52.5 & 79.5 & 86.6 & 478.7 \\
			BFAN* + AOQ \cite{chen2020adaptive} & \textit{ECCV} 2020 & 73.2 & \textbf{94.5} & \textbf{97.0} & 54.0 & 80.3 & \textbf{87.7} & 486.7 \\
			BFAN* + Meta-SPN \cite{wei2021meta} & \textit{ACM MM} 2021 & 72.5 & 93.2 & 96.7 & 53.3 & 80.2 & 87.2 & 483.1 \\
			\rowcolor{black!10}
			\textbf{BFAN*} + \textbf{Unified Loss $\mathcal{L}_{\text{Unified}}$} & \textbf{\textit{Ours}} & \textbf{74.3} &  93.8  & 96.7 & \textbf{54.5} & \textbf{80.8}  & 87.5 & \textbf{487.6}  \\
			
			\hline
			\specialrule{0em}{2pt}{0pt}
			SGRAF* + Triplet-HN \cite{diao2021similarity} & \textit{AAAI} 2021 & 77.8 & 94.1 & 97.4 & 58.5 & 83.0 & 88.8 & 499.6 \\
			SGRAF* + NCR \cite{huang2021learning} & \textit{NeurIPS} 2021 & 77.3 & 94.0 & \textbf{97.5} & 59.6 & 84.4 & 89.9 & 502.7 \\
			\rowcolor{black!10}
			\textbf{SGRAF*} + \textbf{Unified Loss $\mathcal{L}_{\text{Unified}}$} & \textbf{\textit{Ours}} & \textbf{78.3} & \textbf{95.0} & 97.4 & \textbf{60.4} & \textbf{85.0} & \textbf{90.6} & \textbf{506.6} \\
			
			\bottomrule[1pt]
		\end{tabular}
	\end{center}
	\label{other_all}
\end{table*}
\section{Experiments}
\subsection{Implementation Details}
Our all experiments are conducted on an NVIDIA GeForce RTX 3090 GPU using PyTorch.
\subsubsection{Image-Text Retrieval Without Pre-training}
In order to justify the superiority  of our unified loss over the state-of-the-art image-text retrieval models, we conduct experiments on VSE++, BFAN \cite{liu2019focus} and SGRAF \cite{diao2021similarity} by only replacing the loss functions.
\begin{itemize}
	\item \textbf{VSE++} is the most representative algorithm in image-text retrieval and several variant methods have been proposed based on this method.
	We re-implement VSE++ using Bottom-Up and Top-Down (BUTD) attention region features \cite{anderson2018bottom} as our baseline, denoted as VSE++ (BUTD). 
	Image features are extracted by a pretrained Faster R-CNN \cite{anderson2018bottom} with a ResNet-101 \cite{he2016deep} backbone. 
	Image features are mapped to a set of 1024-dimensional vectors through a fully connected layer. 
	This set of vectors is aggregated into a vector through a max pooling layer as the final image embedding.
	A GRU is used to encode each sentence into a text embedding of dimension 1024.
	Our reproduced VSE++ (BUTD) has superior performance compared to the original VSE++ \cite{faghri2018vse++}.
	VSE++ (BUTD) is trained using Adam \cite{kingma2015adam} for 20 epochs, with a batch size of 128 for both datasets. 
	The initial learning rate of the model is set as 0.0005 for first 10 epochs, and then decays by a factor of 10 for the last 10 epochs. 
	Hyper-parameters are set as $ m = 0.2 $ and $ \gamma = 60 $, for both Flickr30K and MS-COCO.
	\item \textbf{BFAN} is a bidirectional focal attention network for image-text matching, which aligns the image and text by focusing on relevant fragments.
	BFAN is trained using Adam for 30 epochs, with a batch size of 128 for both datasets. 
	The initial learning rate of the model is set as 0.0005 for first 15 epochs, and then decays by a factor of 10 for the last 15 epochs. 
	Hyper-parameters are set as $ m = 0.2 $ and $ \gamma = 50 $, for both Flickr30K and MS-COCO.
	\item 
	\textbf{SGRAF} is a similarity graph reasoning and attention filtration network for image-text matching. 
	SGRAF, one of the open-sourced state-of-the-art models, is trained using Adam for 30 epochs, with a batch size of 128 for both datasets. 
	The initial learning rate of the model is set as 0.0005 for first 15 epochs, and then decays by a factor of 10 for the last 15 epochs. 
	Hyper-parameters are set as $ m = 0.2 $ and $ \gamma = 50 $, for both Flickr30K and MS-COCO.
\end{itemize}

\subsubsection{Fine-tuning on Vision-Language Pre-training Model}
In order to verify the improvement of our unified loss on the vision-language pre-training model, we also conduct experiments on CLIP \cite{radford2021learning}.
CLIP is a neural network trained on a variety of image-text pairs. 
It can be instructed in natural language to predict the most relevant text snippet, given an image, without directly optimizing for the task.
We use the official ViT-L/14@336px model for experiments.
We add an embedding layer of dimension 768 after each of the image and text encoders for fine-tuning.
CLIP is fine-tuned using Adam \cite{kingma2015adam} for 10 epochs, with a batch size of 128 for both datasets. 
The learning rate of the model is set as 0.0005. 
Hyper-parameters are set as $ m = 0.2 $ and $ \gamma = 50 $, for both Flickr30K and MS-COCO.

\subsubsection{Comparison to the Other Loss Functions for Vision-Language Retrieval}
There are a number of loss functions proposed for vision-language retrieval, so we compare our unified loss with these losses:
\begin{itemize}
	\item \textbf{SSP}: Self-similarity polynomial loss \cite{wei2020universal} is a weighted triplet loss that defines a weight function for the positive and negative pairs respectively.
	\item \textbf{RSP}: Relative-similarity polynomial loss \cite{wei2021universal} is a weighted triplet loss that assigns appropriate weights to the relative-similarity scores between positive and negative pairs.
	\item \textbf{Meta-SPN}: Meta-SPN \cite{wei2021meta} is a Meta Self-Paced Network that automatically learns a weighting scheme from data for cross-modal matching.
	\item \textbf{AOQ}: Adaptive offline quintuplet loss \cite{chen2020adaptive} provides offline negatives by sampling negatives offline from the whole training set.
	\item \textbf{NCR}: Noisy correspondence rectifier \cite{huang2021learning} is a method for learning with noisy correspondence for cross-modal matching.
\end{itemize}
For a fair comparison, we use the same image-text retrieval model as in the papers of various loss functions above, but simply replace the loss function with our unified loss.
We conduct experiments on SCAN \cite{lee2018stacked}, BFAN \cite{liu2019focus} and SGRAF \cite{diao2021similarity} by only replacing the loss functions.
\begin{itemize}
	\item \textbf{SCAN} is a stacked cross attention network, which measures the image-text similarity by aligning image regions and words.
	SCAN is trained using Adam for 30 epochs, with a batch size of 128 for both datasets. 
	The initial learning rate of the model is set as 0.0005 for first 15 epochs, and then decays by a factor of 10 for the last 15 epochs. 
	Hyper-parameters are set as $ m = 0.2 $ and $ \gamma = 50 $.
	\item \textbf{BFAN} is trained using Adam for 30 epochs, with a batch size of 128 for both datasets. 
	The initial learning rate of the model is set as 0.0005 for first 15 epochs, and then decays by a factor of 10 for the last 15 epochs. 
	Hyper-parameters are set as $ m = 0.2 $ and $ \gamma = 50 $, for both Flickr30K and MS-COCO.
	\item 
	\textbf{SGRAF} is trained using Adam for 30 epochs, with a batch size of 128 for both datasets. 
	The initial learning rate of the model is set as 0.0005 for first 15 epochs, and then decays by a factor of 10 for the last 15 epochs. 
	Hyper-parameters are set as $ m = 0.2 $ and $ \gamma = 50 $.
\end{itemize}

\subsubsection{Video-text retrieval}
In order to verify the improvement of our unified loss on the state-of-the-art video-text retrieval models, we conduct experiments on MMT \cite{gabeur2020multi} and CE \cite{liu2019use} by only replacing the loss functions.
\begin{itemize}
	\item MMT is a multi-modal transformer that  jointly encodes the different modalities in video. 
	The transformer architecture is also leveraged to encode and model the temporal information.
	\item CE is a collaborative experts model that  aggregates information from these different pre-trained experts.
\end{itemize}
Hyper-parameters are set as $ m = 0.2 $ and $ \gamma = 50 $, for both two models.

\subsection{Experimental Results}
\subsubsection{Image-Text Retrieval Without Pre-training}
In the appendix, we compare our method with more image-text retrieval methods.
\tablename~\ref{itr_full} compares our unified loss with image-text retrieval baselines on Flickr30K and MS-COCO 1K test set.
For a fair comparison, the feature extraction backbone of all methods is the same, \textit{i.e.},  for image is Faster R-CNN \cite{anderson2018bottom}, and  for text is Bi-directional GRU (Bi-GRU) \cite{schuster1997bidirectional}.
Most methods use Triplet-HN as the optimization target. 
By replacing the loss function with our unified loss, the performance of the three baseline methods is improved.
On the Flickr30K dataset, \textbf{SGRAF* + $\mathcal{L}_{\text{Unified}}$} improves RSUM by 7.0\% compared to SGRAF*. 
On MS-COCO 1K test set, applying our unified loss to SGRAF* can improve RSUM by 3.1\%.
\tablename~\ref{itr_coco5k} summarizes the retrieval results on MS-COCO 5K test set.
Unified loss boosts the performance of almost all evaluation metrics on the three baselines on the MS-COCO 5K test set.
Our unified loss yields a 1.4\% increase for image-to-text retrieval and 1.3\% improvement for text-to-image retrieval in terms of R@1 when compared with SGRAF*.

\subsubsection{Comparison to the Other Loss Functions for Vision-Language Retrieval}
In the appendix, we compare the performance of different loss functions on more image-text retrieval methods.
For a fair comparison, we use the same image-text retrieval model as in the papers of various loss functions above, but simply replace the loss function with our unified loss.
The experimental results of unified loss compared to the other losses on Flickr30K are shown in \tablename~\ref{other_all}.
Compared to the original Triplet-HN, our unified loss achieves a substantial performance gain. 
Compared with other loss functions, unified loss has also improved most of the evaluation metrics.
Unified loss does not need to introduce many hyperparameters to assign weights like SSP and RSP, nor does it need to train an additional network for weight assignment like Meta-SPN.
Compared with AOQ, Unified loss only requires online hard negative mining, and does not increase the complexity.

\clearpage

{\small
	\bibliographystyle{ieee_fullname}
	\bibliography{egbib}
}

\end{document}